\pdfoutput=1

\documentclass[11pt]{article}

\usepackage[]{ACL2023}

\usepackage{times}
\usepackage{latexsym}
\usepackage{enumitem}
\usepackage{graphicx}
\usepackage{subfigure}
\usepackage{multirow}
\usepackage{amsmath}
\usepackage{booktabs} 
\usepackage{amssymb}
\usepackage[T1]{fontenc}

\usepackage[utf8]{inputenc}

\usepackage{microtype}

\usepackage{inconsolata}

\newcommand{\model}{MED}

\title{Croppable Knowledge Graph Embedding}

\author{Yushan Zhu$^{\spadesuit\diamondsuit}$, Wen Zhang$^{\spadesuit\diamondsuit}$, Zhiqiang Liu$^{\spadesuit\diamondsuit}$, Mingyang Chen$^\spadesuit$, \\ {\bf Lei Liang$^\clubsuit$, Huajun Chen$^{\spadesuit\diamondsuit\heartsuit\ast}$} \\
     $^\spadesuit$ Zhejiang University  $^\clubsuit$Ant Group \\ 
     $^\diamondsuit$ Zhejiang University - Ant Group Joint Laboratory of Knowledge Graph \\ 
     $^\heartsuit$ Zhejiang Key Laboratory of Big Data Intelligent Computing \\
     \texttt{\{yushanzhu, zhang.wen, zhiqiangliu, mingyangchen, huajunsir\}@zju.edu.cn}}

\begin{document}
\maketitle
\begin{abstract}
Knowledge Graph Embedding (KGE) is a common approach for Knowledge Graphs (KGs) in AI tasks. Embedding dimensions depend on application scenarios. Requiring a new dimension means training a new KGE model from scratch, increasing cost and limiting efficiency and flexibility. 
In this work, we propose a novel KGE training framework MED. It allows one training to obtain a croppable KGE model for multiple scenarios with different dimensional needs. Sub-models of required dimensions can be directly cropped and used without extra training. In MED, we propose a mutual learning mechanism to improve the low-dimensional sub-models and make high-dimensional sub-models retain the low-dimensional sub-models' capacity, an evolutionary improvement mechanism to promote the high-dimensional sub-models to master the triple that the low-dimensional sub-models can not, and a dynamic loss weight to adaptively balance the multiple losses. Experiments on 4 KGE models across 4 standard KG completion datasets, 3 real-world scenarios using a large-scale KG, and extending MED to the BERT language model demonstrate its effectiveness, high efficiency, and flexible extensibility. 

\end{abstract}

\section{Introduction}
Knowledge Graphs (KGs) consist of triples in the form of (\emph{head entity, relation, tail entity}), abbreviated as (\emph{h, r, t}). KGs are widely used in recommendation systems~\cite{DBLP:conf/mm/ZhuZZYCZC21}, information extraction~\cite{DBLP:conf/i-semantics/DaiberJHM13}, question answering~\cite{DBLP:conf/www/DiefenbachSM18}, etc. A common KG application way is knowledge graph embedding (KGE) \cite{DBLP:conf/nips/BordesUGWY13, DBLP:conf/iclr/SunDNT19}, which maps KG entities and relations into continuous vector spaces for various tasks.

\begin{figure}[t]
\vspace{-5mm}
	\centering 
\setlength{\abovecaptionskip}{-0.cm}
\setlength{\belowcaptionskip}{-0.cm}
  \includegraphics[width=1.0\linewidth]{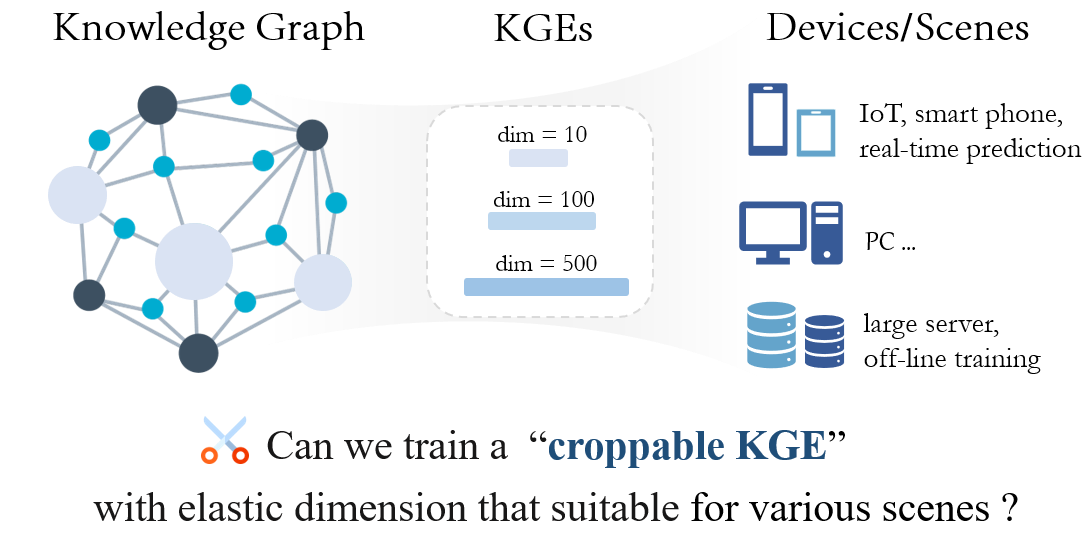}
\caption{Diverse KGE dimensions for a KG.}
    \label{fig:intro}
    \vspace{-5mm}
\end{figure}

Higher-dimensional KGEs have more expressive power and better performance but come with more parameters, demanding more storage and computing resources \cite{DBLP:conf/wsdm/ZhuZCCC0C22,DBLP:conf/acl/Sachan20}.
The suitable KGE dimensions vary by device or scenario. As Fig.~\ref{fig:intro} shows, large remote servers can support high-dimensional KGEs, while small-to-medium terminal devices like in-vehicle systems or smartphones can only handle low-dimensional ones due to resource limits.
Thus, people prefer training high-quality KGEs with appropriate dimensions. However, a new KGE must be trained from scratch when a new dimension is needed. Especially for low-dimensional KGEs, model compression techniques like knowledge distillation \cite{DBLP:journals/corr/HintonVD15,DBLP:conf/wsdm/ZhuZCCC0C22} are required to ensure performance. This raises training costs and restricts KGE's efficiency and flexibility across scenarios.

Thus a new concept "\underline{croppable KGE}" is proposed and we are interested in the research question that 
\textbf{is it possible to train a croppable KGE, with which KGEs of various required dimensions can be cropped out of it and directly used without extra training, and perform well?} 

In this work, our croppable KGE learning idea is to train an entire KGE with multiple different-dimensional sub-models sharing embedding parameters and trained simultaneously. 
The aim is for low-dimensional sub-models to benefit from high-dimensional ones, and high-dimensional sub-models to retain low-dimensional ones' ability and learn what they can't.
Based on this, we propose a croppable KGE training framework \textbf{{\model}}, with three main modules: \underline{\textbf{M}}utual learning mechanism, \underline{\textbf{E}}volutionary improvement mechanism, and \underline{\textbf{D}}ynamic loss weight.
Specifically, the \textbf{mutual learning mechanism}, based on knowledge distillation, enables neighbor sub-models to learn from each other, improving low-dimensional sub-model performance and helping high-dimensional ones retain low-dimensional abilities. The \textbf{evolutionary improvement mechanism} helps high-dimensional focus on and master triples that low-dimensional ones can't correctly predict. The \textbf{dynamic loss weight} adaptively balances sub-models' losses according to dimension to enhance overall performance.

We evaluate the effectiveness of {\model} by implementing it on four typical KGE methods and four standard KG datasets. We prove its practical value via a real-world large-scale KG and downstream tasks, and demonstrate its extensibility on BERT \cite{DBLP:conf/naacl/DevlinCLT19} and GLUE \cite{DBLP:conf/iclr/WangSMHLB19} benchmarks. 
Experimental results show that: (1) {\model} trains a croppable KGE for various dimensional needs, with high-performing, parameter-shared sub-models ready for direct use. (2) {\model} has far higher training efficiency than independent training or knowledge distillation. (3) {\model} can be extended to other neural networks and performs well. (4) The three modules in {\model} are crucial for optimal performance.
In summary, our contributions are as follows:
\begin{itemize}
\item We introduce the research question and task of training croppable KGE, allowing direct use of differently-dimensional KGEs.
\item We propose a novel framework {\model}, including a mutual learning mechanism, an evolutionary improvement mechanism, and a dynamic loss weight, to ensure the overall performance of all sub-models. 
\item We show that {\model}'s sub-models perform well, with low-dimensional ones outperforming KGEs trained by SOTA distillation methods. {\model} also shows good performance and extensibility in real-world applications and other types of neural networks.
\end{itemize}

\section{Related Work}
This work is to achieve a croppable KGE for different dimensional needs. A common way to get a good-performance KGE of the target dimension is using knowledge distillation with a high-dimensional powerful teacher KGE. So, we focus on two relevant research areas: knowledge graph embedding and knowledge distillation.
\subsection{Knowledge Graph Embedding}
Knowledge graph embedding (KGE) maps KG entities and relations into continuous vector spaces for downstream tasks. 
TransE~\cite{DBLP:conf/nips/BordesUGWY13}, a representative translation-based KGE, treats relations as translations between entities. Its variants include TransH\cite{DBLP:conf/aaai/WangZFC14}, TransR\cite{DBLP:conf/aaai/LinLSLZ15}, TransD\cite{DBLP:conf/acl/JiHXL015}, etc.
RESCAL\cite{DBLP:conf/icml/NickelTK11}, based on vector decomposition, was followed by improvements like DistMult\cite{DBLP:journals/corr/YangYHGD14a}, ComplEx\cite{DBLP:conf/icml/TrouillonWRGB16}, and SimplE\cite{DBLP:conf/nips/Kazemi018}.
RotatE\cite{DBLP:conf/iclr/SunDNT19}, a rotation-based method, views relations as rotations between entities, similar to QuatE\cite{DBLP:conf/nips/0007TYL19} and DihEdral\cite{DBLP:conf/acl/XuL19}. 
PairRE\cite{DBLP:conf/acl/ChaoHWC20} uses two relation vectors for complex pattern encoding.
With neural network development, KGEs based on graph neural networks (GNNs)~\cite{DBLP:conf/aaai/DettmersMS018,DBLP:conf/naacl/NguyenNNP18,DBLP:conf/esws/SchlichtkrullKB18,DBLP:conf/iclr/VashishthSNT20} also emerged. 
While KGEs are simple and effective, a key challenge remains: Different scenarios demand different KGE dimensions based on device resources. Training a new KGE model from scratch for each dimension requirement hikes training costs and restricts flexibility in serving diverse scenarios.
\subsection{Knowledge Distillation}
High-dimensional KGEs possess strong expressive power thanks to numerous parameters. However, they demand substantial storage and computing resources, rendering them unsuitable for all scenarios, particularly those involving small devices. To address this, a prevalent approach is to compress a high-dimensional KGE into target dimension via knowledge distillation \cite{DBLP:journals/corr/HintonVD15,DBLP:conf/aaai/MirzadehFLLMG20} and quantization~\cite{DBLP:conf/acl/BaiZHSJJLLK20,DBLP:conf/iclr/StockFGGGJJ21} technology.

Quantization replaces continuous vector representations with lower-dimensional discrete codes. TS-CL~\cite{DBLP:conf/acl/Sachan20} was the first to apply quantization in KGE compression.
LightKG~\cite{DBLP:conf/cikm/WangWLG21}
uses a residual module to create diversity in codebooks.
Yet, as quantization doesn't boost inference speed, it's still not suitable for devices with limited computing resources.

Knowledge distillation (KD) is widely used in Computer Vision~\cite{DBLP:conf/aaai/MirzadehFLLMG20} and Natural Language Processing~\cite{DBLP:conf/naacl/DevlinCLT19,DBLP:conf/emnlp/SunCGL19} to shrink model size and boost inference speed. Its core is using a large teacher model's output to guide a small student model's training.
DualDE \cite{DBLP:conf/wsdm/ZhuZCCC0C22} is a representative KD-based work to transfer knowledge from high- to low-dimensional KGE, considering teacher-student mutual influences. 
MulDE~\cite{DBLP:conf/www/Wang0MS21} transfers knowledge from multiple hyperbolic KGE models to one student. ISD~\cite{DBLP:journals/corr/abs-2206-02963} improves low-dimensional KGE by playing teacher and student roles alternatively. IterDE~\cite{DBLP:conf/aaai/LiuWSW23} introduces iterative distillation for smooth knowledge transfer. 
Other KG-related distillation works include PMD~\cite{DBLP:conf/aaai/FanCXK0L24}, which applies distillation to pre-trained language models for KG completion, IncDE~\cite{DBLP:conf/aaai/LiuK0SGLJL24}, using distillation between same-dimensional models at different times for incremental learning, and SKDE~\cite{DBLP:conf/coling/XuWF24}, which proposes self-knowledge distillation to avoid a complex teacher model.
Among these, DualDE~\cite{DBLP:conf/wsdm/ZhuZCCC0C22} and IterDE~\cite{DBLP:conf/aaai/LiuWSW23} are most relevant to our work as they compress high-dimensional teacher into low-dimensional student. 
In this work, we propose a novel KD-based KGE training framework {\model}, one training can obtain a croppable KGE that meets multiple dimensional requirements.

\section{Preliminary}
\begin{table}[ht]
\setlength\tabcolsep{3pt}
\vspace{-0.3cm}
\setlength{\abovecaptionskip}{0.1cm}
\setlength{\belowcaptionskip}{-0.1cm}
\centering
\resizebox{0.45\textwidth}{!}{
\begin{tabular}{c|c}
\toprule
KGE method & Scoring Function $f(\mathbf{h}, \mathbf{r}, \mathbf{t})$\\
\midrule
TransE~\cite{DBLP:conf/nips/BordesUGWY13} & $-\left \|\mathbf{h}+\mathbf{r}-\mathbf{t}\right \|$   \\

SimplE~\cite{DBLP:conf/nips/Kazemi018} & $ \frac{1}{2}(<h^H, r, t^T> + <t^H, r^{-1}, h^T>)$\\
RotatE~\cite{DBLP:conf/iclr/SunDNT19}  & $-\left \|  \mathbf{h}\circ \mathbf{r} - \mathbf{t} \right \|$   \\

PairRE~\cite{DBLP:conf/acl/ChaoHWC20}  & $-\left \| \mathbf{h}\circ \mathbf{r}^H-\mathbf{t}\circ \mathbf{r}^T \right \|$ \\
\bottomrule
\end{tabular}
}
\caption{Score functions.}
\label{t_score}
\vspace{-2mm}
\end{table}
Knowledge graph embedding (KGE) methods use a scoring function $f$ to represent entities and relations in a continuous vector space. Given a KG $\mathcal{G}=(\mathcal{E}, \mathcal{R}, \mathcal{T})$ where $\mathcal{E}$, $\mathcal{R}$ and $\mathcal{T}$ are sets of entities, relations and observed triples,  for a triple $(h,r,t)$ ($h\in \mathcal{E}, r\in \mathcal{R}$, $t\in \mathcal{E}$), we use the triple scoring function $s_{(h,r,t)} = f(\mathbf{h},\mathbf{r},\mathbf{t})$ (with entity and relation embeddings as input) to measure triple plausibility in the embedding space. 
Table~\ref{t_score} summarizes scoring functions of some popular KGE methods, where $\circ$ is the Hadamard product, $ <x^1,..., x^k> = \sum_i x^1_i...x_i^k $ is the generalized dot product. Higher triple scores mean the model is more likely to consider triples true. The optimization objective of the KGE model is
\begin{equation}
\begin{aligned}
    L_{KGE} = - \sum_{(h,r,t)\in \mathcal{T}\cup \mathcal{T}^-} y \log \sigma(s_{(h, r, t)}) 
\\ + (1-y)\log (1-\sigma(s_{(h, r, t)})),
\end{aligned}
\end{equation}
where $\mathcal{T}^-= \mathcal{E} \times \mathcal{R} \times \mathcal{E} \setminus \mathcal{T}$ is negative triple set, $\sigma$ is Sigmoid function, $y$ is ground-truth label of $(h,r,t)$, $y=1$ for positive and $0$ for negative one.

\section{{\model} Framework}
\begin{figure}
	\centering 
\vspace{-0.1cm}
\setlength{\abovecaptionskip}{0.1cm}
\setlength{\belowcaptionskip}{-0.3cm}
\includegraphics[width=0.8\linewidth]{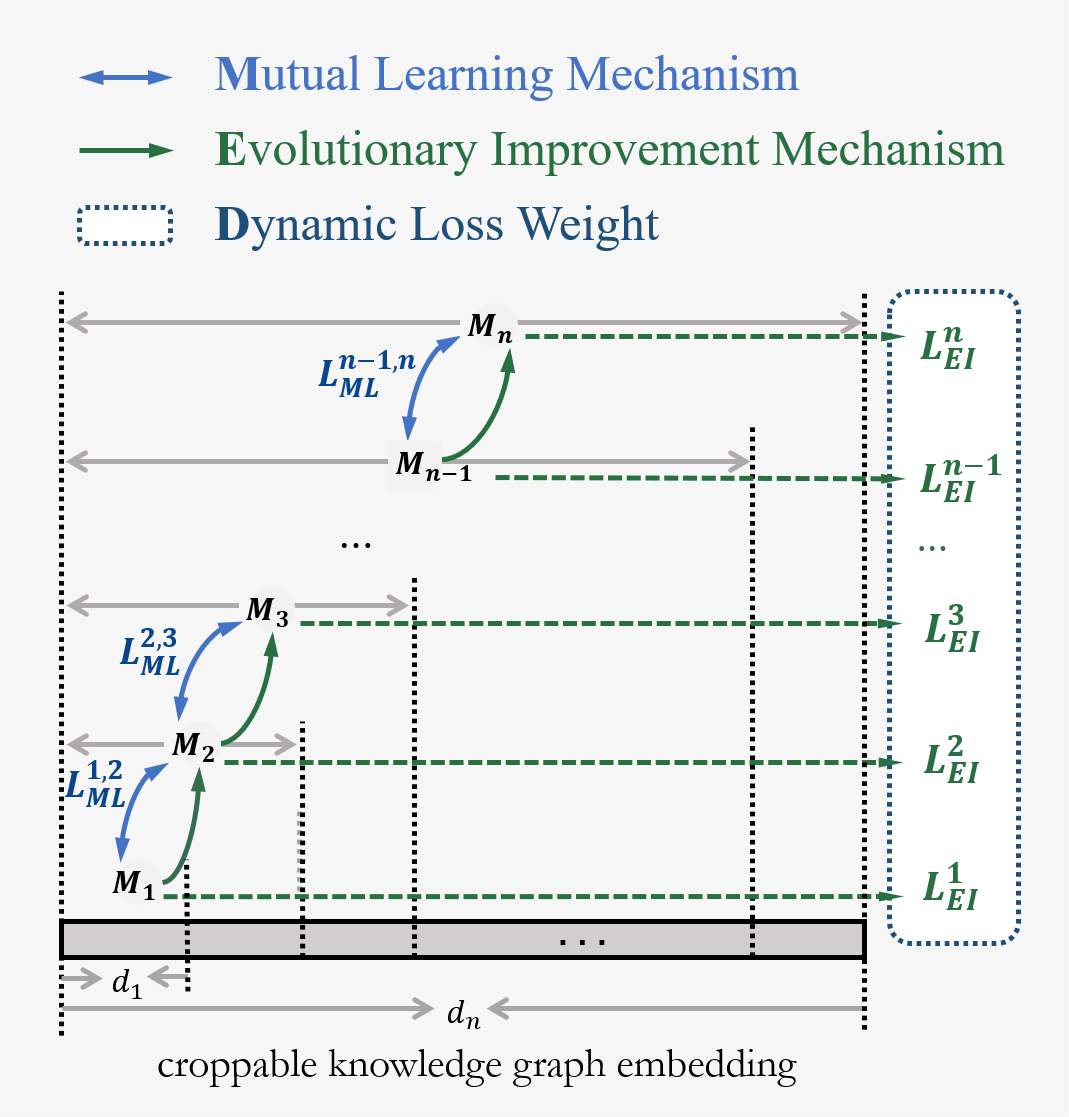}
\caption{Overview of {\model}.}
    \label{fig:model}
\end{figure}
As in Fig.~\ref{fig:model}, our croppable KGE framework {\model}
contains $n$ sub-models of different dimensions, $M_i (i=1,2...,n)$ with dimension of $d_i$. Each $M_i$ consists of the first $d_i$ dimensions of the full embedding. The triple $(h,r,t)$ score from $M_i$ is $s_{(h,r,t)}^i=f(\mathbf{h}[0$$:$$d_i], \mathbf{r}[0$$:$$d_i], \mathbf{t}[0$$:$$d_i])$, where $\mathbf{h}[0$$:$$d_i]$ are the first $d_i$ elements of vector $\mathbf{h}$. 
The parameters of sub-model $M_i$ are shared by higher-dimensional $M_j(i$$<$$j$$\leqslant$$n)$.
The number of sub-models $n$ and each sub-model's dimension $d_i$ can be set per actual needs. 
For low-dimensional sub-models, we aim to maximize performance. 
High-dimensional sub-models should not only replicate the capabilities of low-dimensional ones but also learn what low-dimensional ones can't, that is, correctly predicting triples mispredicted by low-dimensional sub-models.
{\model} is based on knowledge distillation~\cite{DBLP:journals/corr/HintonVD15,DBLP:journals/corr/abs-1903-12136,DBLP:conf/naacl/DevlinCLT19} technique, where student fits hard (ground-truth) label and soft label from teacher simultaneously. 
In {\model}, we first introduce a \textit{mutual learning mechanism}. This helps low-dimensional sub-models learn from high-dimensional ones for better performance, and vice versa, so high-dimensional sub-models retain the capabilities of low-dimensional ones.
Next, we present an \textit{evolutionary improvement mechanism} to let high-dimensional sub-models acquire knowledge that low-dimensional ones can't learn well.  Finally, we train {\model} with \textit{dynamic loss weight} to adaptively balance multiple optimization goals of sub-models.

\subsection{Mutual Learning Mechanism}
\label{sec:MLM}
We consider each sub-model $M_i$ as the student of its higher-dimensional neighbor $M_{i+1}$ for better performance, as high-dimensional KGEs with more parameters are more expressive~\cite{DBLP:conf/acl/Sachan20,DBLP:conf/wsdm/ZhuZCCC0C22}. We also make $M_i$ the student of its lower-dimensional neighbor $M_{i-1}$, enabling the higher-dimensional sub-model to review and retain the lower-dimensional one's capabilities. So, pairwise neighbor sub-models are both teacher and student, learning from each other. Mutual learning loss between each pair of neighbor sub-models is
\begin{equation}
    L_{ML}^{i-1,i}=\sum_{(h, r, t) \in \mathcal{T} \cup \mathcal{T}^-} d_\delta\left(s_{(h, r, t)}^{i-1}, s_{(h, r, t)}^{i}\right), 1< i \leqslant n ,
    \label{equ:Loss_ML}
\end{equation}
where $s_{(h,r,t)}^i$ is the score of triple $(h, r, t)$ from sub-model $M_i$,  indicating the triple's existence likelihood. 
$d_{\delta}$ is Huber loss~\cite{Huber1964Robust} with $\delta=1$, often used in KGE knowledge distillation\cite{DBLP:conf/wsdm/ZhuZCCC0C22}.
In {\model}, each sub-model learns only from its neighbor sub-models. 
This not only cuts down training computational complexity but also keeps the dimension gap between each teacher-student pair relatively small. A large dimension gap between them can undermine the distillation effect \cite{DBLP:conf/aaai/MirzadehFLLMG20,DBLP:conf/wsdm/ZhuZCCC0C22}, making our approach crucial and effective.

\subsection{Evolutionary Improvement Mechanism}
In knowledge distillation \cite{DBLP:journals/corr/HintonVD15}, the hard (ground-truth) label is a key supervision signal during training. 
High-dimensional sub-models need to master triples that low-dimensional sub-models can not, that is correctly predicting positive (negative) triples mispredicted as negative (positive) by low-dimensional sub-models.
In {\model}, for a triple $(h, r, t)$, sub-model $M_i$'s optimization weight for it depends on triple score from $M_{i-1}$. 

For a positive triple, $M_i$'s optimization weight is negatively correlated with the score from $M_{i-1}$.  If $M_{i-1}$ gives a high score (correctly identifying it as positive), $M_i$'s optimization weight for it is low. If $M_{i-1}$ gives a low score (misidentifying it as negative), $M_i$'s optimization weight is high as $M_{i-1}$ can't predict it well.
The optimization weight of $M_i$ for the positive triple is
\begin{align}
\label{equ:pos_w}
\begin{split}
pos_{h,r,t}^i=
\left\{
\begin{array}{ll}
 1 / |T_{batch}| ,& i = 1
\\
 \frac{\exp w_1 / s_{(h, r, t)}^{i-1}}{\sum\limits_{(h,r,t)\in \mathcal{T}_{b}} \exp w_1 / s_{(h, r, t)}^{i-1}} , & 1< i \leqslant n,
\end{array}
\right.
\end{split}
\end{align}
where $s_{(h,r,t)}^{i-1}$ is the score for triple $(h, r, t)$ from $M_{i-1}$, $\mathcal{T}_{b}$ is positive triple set in a batch, and $w_1$ is a learnable scaling parameter.
Conversely, for a negative triple, $M_i$'s optimization weight is positively correlated with the score from $M_{i-1}$, as also used in ~\cite{DBLP:conf/iclr/SunDNT19}. The optimization weight of $M_i$ for negative triple is
\begin{align}
\label{equ:neg_w}
\begin{split}
neg_{h,r,t}^i=
\left\{
\begin{array}{ll}
 1 / |T_{batch}^-|  , & i = 1\\
 \frac{\exp w_2 \cdot s_{(h, r, t)}^{i-1}}{\sum\limits_{(h,r,t)\in \mathcal{T}_{b}^{-}} \exp w_2 \cdot s_{(h, r, t)}^{i-1}}  , & 1< i \leqslant n,
\end{array}
\right.
\end{split}
\end{align}
where $\mathcal{T}_{b}^-$ is negative triple set in a batch, and $w_2$ is a learnable scaling parameter. The evolutionary improvement loss of the sub-model $M_i$ is
\begin{equation}
\begin{aligned}
\label{equ:loss_EI}
    L_{EI}^i = - \sum_{(h,r,t)\in \mathcal{T}\cup \mathcal{T}^-}
 pos_{h, r, t}^i \cdot  y \log \sigma(s_{(h, r, t)}^i) \\
+ neg_{h, r, t}^i\cdot   (1-y)\log (1-\sigma(s_{(h, r, t)}^i)),
\end{aligned}
\end{equation}
In each sub-model, different hard-label loss weights are set for different triples, and high-dimensional sub-models focus more on triples low-dimensional ones can't learn well.

\subsection{Dynamic Loss Weight}
As {\model} optimizes multiple sub-models, we use dynamic loss weights during training. 
At first, low-dimensional sub-models prioritize learning from high-dimensional ones to improve performance, relying more on soft label information. Thus, for them, evolutionary improvement loss should be less than mutual learning loss.
Conversely, high-dimensional sub-models should focus more on what low-dimensional ones can't correctly predict and reduce the impact of low-quality low-dimensional outputs. They rely more on hard label information, so their evolutionary improvement loss should exceed mutual learning loss. 
For a teacher-student pair, the mutual learning loss affects both models equally. We set different evolutionary improvement loss weights for sub-models, and the final training loss function of {\model} is
\begin{equation}
    L = \sum_{i=2}^n L_{ML}^{i-1,i} 
 + \sum_{i=1}^n \exp (\frac{w_3 \cdot d_i}{d_n}) \cdot L_{EI}^i,
    \label{equ:loss}
\end{equation}
where $w_3$ is a learnable scaling parameter, and $d_i$ is the dimension of the $i$th sub-model.

\section{Experiment}
\label{sec:experiment}
We evaluate {\model} on typical KGE and GLUE benchmarks and address these research questions: (\textbf{RQ1}) Can {\model} train a croppable KGE at once, with multiple differently-dimensional sub-models cropped from it, all performing well? (\textbf{RQ2}) Can MED yield parameter-efficient KGE models? (\textbf{RQ3}) Does MED work in real-world applications? (\textbf{RQ4}) Can MED be extended to non-KGE neural networks?


\subsection{Experiment Setting}
\label{sec:exper_set}

\subsubsection{Dataset and KGE methods}

{\model} is universal and applicable to any KGE method with a triple score function. We take 4 common KGE methods in Table~\ref{t_score} as examples: TransE~\cite{DBLP:conf/nips/BordesUGWY13}, SimplE~\cite{DBLP:conf/nips/Kazemi018}, RotatE~\cite{DBLP:conf/iclr/SunDNT19} and PairRE~\cite{DBLP:conf/acl/ChaoHWC20}. 

\begin{table}[ht]
\setlength\tabcolsep{5pt}
\vspace{-0.1cm}
\setlength{\abovecaptionskip}{0cm}
\setlength{\belowcaptionskip}{-0.2cm}
\begin{center}
\resizebox{0.4\textwidth}{!}{
\begin{tabular}{lccccc}
\toprule
Dataset   & \#Ent. & \#Rel. & \#Train & \#Valid &\#Test \\
\midrule
WN18RR    & 40,943      & 11         & 86,835  & 3,034   & 3,134  \\
FB15K237 & 14,541      & 237        & 272,115 & 17,535  & 20,466 \\
CoDEx-L & 77,951 & 69 & 551,193 & 30,622 &30,622 \\
YAGO3-10 & 123,143 & 37 & 1,079,040 & 4,978 &4,982 \\
SKG & 6,974,959 & 15 & 50,775,620 & - & - \\
\bottomrule
\end{tabular}
}
\end{center}
\caption{Statistics of datasets.}
\label{table_dataset}
\vspace{-0.2cm}
\end{table}
We conduct comparison experiments on on four KG completion benchmark datasets: two common ones, WN18RR \cite{DBLP:conf/emnlp/ToutanovaCPPCG15} and FB15K237 \cite{DBLP:conf/aaai/DettmersMS018}, and two larger-scale ones, CoDEx-L \cite{DBLP:conf/emnlp/SafaviK20} and YAGO3-10 \cite{DBLP:conf/cidr/MahdisoltaniBS15}. 
In real-world scenarios, we apply {\model} to a large-scale e-commerce social knowledge graph (SKG) from Taobao, which has over 50 million social record triples from about 7 million users. Table~\ref{table_dataset} presents dataset statistics.

\subsubsection{Evaluation Metric}
For link prediction task, we adopt standard metrics MRR and Hit@$k$ $(k=1,3,10)$ 
with filter setting \cite{DBLP:conf/nips/BordesUGWY13}. For a test triple $(h, r, t)$, we construct candidate triples by replacing $h$ with all entities and calculate the triple score rank of $(h, r, t)$ among candidate triples as the head prediction rank $rank_{h}$. Likewise, we get the tail prediction rank $rank_{t}$. We average $rank_{h}$ and $rank_{t}$ as $(h, r, t)$'s final rank. MRR is the mean reciprocal rank of all test triples, Hit@$k$ is the percentage of test triples with rank $\le k$.  
We use \textit{Effi}=\textit{MRR/\#P} (\textit{\#P} is the parameter number)~\cite{DBLP:conf/aaai/ChenZYZGPC23} to measure models' parameter efficiency.
We use f1-score and accuracy for user labeling task, and normalized discounted cumulative gain ndcg@$k (k=5,10)$ for product recommendation task.

\subsubsection{Implementation}
\label{sec:implementaion}
For the link prediction task, 
we set $d_n=640$ for the highest-dimensional sub-model $M_n$ and $d_1=10$ for the lowest-dimensional sub-model $M_1$. We set $n=64$ and the dimension gap $10$ for every pair of neighbor sub-models. There are a total of $64$ available sub-models of different dimensions from 10 to 640 in our croppable KGE model. The dimension of $M_i (i=1,2...,64)$ is $10\times i$. For the user labeling and product recommendation task, we set $n=3$ and train the croppable KGE containing 3 sub-models: $M_1$ with $d_1=10$ for mobile phone (MB) terminals that are limited by storage and computing resources, $M_2$ with $d_2=100$ for the personal computer (PC), and $M_3$ with $d_3=500$ for the platform's servers. We initialize the learnable scaling parameters $w_1, w_2$ and $w_3$ in \eqref{equ:pos_w}, \eqref{equ:neg_w} and \eqref{equ:loss} to 1.
We implement {\model} by extending OpenKE \cite{DBLP:conf/emnlp/HanCLLLSL18}, an open-source KGE framework based on PyTorch. 
We set the batch size to $1024$ and the maximum training epoch to $3000$ with early stopping. 
For each positive triple, we generate $64$ negative triples by randomly replacing its head or tail entity with another entity. We use Adam~\cite{DBLP:journals/corr/KingmaB14} optimizer with a linear decay learning rate scheduler and perform a search on the initial learning rate in $\{0.0001,0.0005,0.001,0.01\}$. We train all sub-models simultaneously by optimizing the uniformly sampled sub-models from the full Croppable model in each step.

\subsubsection{Baselines}
\label{sec:baseline}
For each required dimension $d_r$, we extract the first $d_r$ dimensions from our croppable KGE as the target model and compare it to the KGE models obtained by 8 baselines of the following 3 types: 
\begin{itemize}[leftmargin=*]
    \item Directly training the target KGE model of requirement dimension $d_r$, referred to as \textbf{1) DT}. The directly trained highest-dimensional KGE model ($d_r=d_n$) is denoted as $M_{max}^{DT}$.
    \item Extracting the first $d_r$ dimensions from $M_{max}^{DT}$ as the target model, called \textbf{2) Ext}. 
    We update $M_{max}^{DT}$ by arranging 640 dimensions in descending order based on their importance before extracting as~\cite{DBLP:conf/iclr/MolchanovTKAK17,DBLP:conf/acl/VoitaTMST19}: \textbf{3) Ext-L}, the importance for each dimension of $M_{max}^{DT}$ is the variation of KGE loss on validation set after removing it; and \textbf{4) Ext-V}, the importance for each dimension is the average absolute of parameter weights of all entities and relations.
    \item Distilling the target KGE by KD methods: \textbf{5) BKD} \cite{DBLP:journals/corr/HintonVD15} is the most basic one by minimizing the KL divergence of the output distributions of teacher and student; \textbf{6) TA}~\cite{DBLP:conf/aaai/MirzadehFLLMG20} uses a medium-size teaching assistant (TA) model as a bridge for size gap, where TA model has the same dimension as the directly trained one whose MRR is closest to the average MRR of teacher and student. We also compare with two KD methods proposed for KGE, which have similar configurations to ours, i.e. compressing high-dimensional teacher into low-dimensional student: \textbf{7) DualDE}~\cite{DBLP:conf/wsdm/ZhuZCCC0C22} considers the mutual influences between teacher and student and optimizes them simultaneously; 
    \textbf{8) IterDE}~\cite{DBLP:conf/aaai/LiuWSW23} enables the KGE model to alternately act as student and teacher so that knowledge can be transferred smoothly between high-dimensional teacher and low-dimensional student.
    In these baselines, $M_{max}^{DT}$ is the teacher, and other settings including hyperparameters are the same as their original papers.
\end{itemize}

\subsection{Performance Comparison}
\label{sec:exper_compare}
We report the link prediction results of some representative dimensions in Table~\ref{table_main_wn_fb}, more results of other dimensions and metrics are in Appendix~\ref{sec:full_result} and the ablation studies are in Appendix~\ref{sec:exper_ablation}.

\begin{table*}[ht]
\vspace{-5mm}
\setlength\tabcolsep{4pt}
\renewcommand\arraystretch{.8}
\setlength{\abovecaptionskip}{-0.cm}
\setlength{\belowcaptionskip}{-0.0cm}
\begin{center}
        \resizebox{.8\textwidth}{!}{
\begin{tabular}{llcc|cc|cc|cclcc|cc|cc|cc}
\toprule
                        &                 & \multicolumn{8}{c}{WN18RR}                                                                                                            &  & \multicolumn{8}{c}{FB15K237}                                                                                                          \\ \cline{3-10} \cline{12-19} 
                        &                 & \multicolumn{2}{c}{10d}         & \multicolumn{2}{c}{40d}         & \multicolumn{2}{c}{160d}        & \multicolumn{2}{c}{640d}        &  & \multicolumn{2}{c}{10d}         & \multicolumn{2}{c}{40d}         & \multicolumn{2}{c}{160d}        & \multicolumn{2}{c}{640d}        \\
\textit{KGE}            & \textit{Method} & \textit{MRR}   & \textit{H10}   & \textit{MRR}   & \textit{H10}   & \textit{MRR}   & \textit{H10}   & \textit{MRR}   & \textit{H10}   &  & \textit{MRR}   & \textit{H10}   & \textit{MRR}   & \textit{H10}   & \textit{MRR}   & \textit{H10}   & \textit{MRR}   & \textit{H10}   \\
\midrule
\multirow{9}{*}{TransE} & DT              & 0.121          & 0.287          & 0.214          & 0.496          & 0.233          & 0.531          & 0.237          & 0.537          &  & 0.150          & 0.235          & 0.299          & 0.477          & 0.315          & 0.499          & 0.322          & \textbf{0.508} \\
                        & Ext             & 0.125          & 0.298          & 0.199          & 0.468          & 0.225          & 0.515          & 0.237          & 0.537          &  & 0.115          & 0.211          & 0.236          & 0.392          & 0.286          & 0.462          & 0.322          & 0.508          \\
                        & Ext-L           & 0.139          & 0.315          & 0.224          & 0.497          & 0.236          & \textbf{0.534} & 0.237          & 0.537          &  & 0.109          & 0.194          & 0.232          & 0.381          & 0.285          & 0.462          & 0.322          & 0.508          \\
                        & Ext-V           & 0.139          & 0.309          & 0.222          & 0.494          & 0.236          & 0.532          & 0.237          & 0.537          &  & 0.139          & 0.256          & 0.237          & 0.396          & 0.293          & 0.466          & 0.322          & 0.508          \\
                        & BKD             & 0.141          & 0.323          & 0.226          & 0.513          & 0.233          & 0.531          & -              & -              &  & 0.176          & 0.293          & 0.303          & 0.480          & 0.315          & 0.501          & -              & -              \\
                        & TA              & 0.144          & 0.335          & 0.226          & 0.512          & 0.234          & 0.533          & -              & -              &  & 0.175          & 0.246          & 0.303          & 0.484          & 0.319          & 0.504          & -              & -              \\
                        & DualDE          & 0.148          & 0.337          & 0.225          & 0.514          & 0.235          & 0.533          & -              & -              &  & 0.179          & 0.301          & 0.306          & 0.483          & 0.319          & 0.505          & -              & -              \\
                        & IterDE  
                        &0.143	&0.332	&0.224	&0.511	&0.236	&0.531	&-	&-	&	&0.176	&0.285	&0.307	&0.482	&0.317	&0.505	&-	&- \\
                        & {\model}        & \textbf{0.170} & \textbf{0.388} & \textbf{0.232} & \textbf{0.518} & \textbf{0.236} & 0.529          & \textbf{0.237} & \textbf{0.537} &  & \textbf{0.196} & \textbf{0.341} & \textbf{0.308} & \textbf{0.486} & \textbf{0.320} & \textbf{0.505} & \textbf{0.322} & 0.507          \\
                        \midrule
\multirow{9}{*}{SimplE} & DT              & 0.061	& 0.126&	0.316&	0.389&	0.409&	0.459&	0.421&	0.481	&	&0.097&	0.179	&0.236&	0.390&	0.285	&0.458&	\textbf{0.295}&	\textbf{0.472} \\
                        & Ext             & 0.004	&0.007	&0.160	&0.249	&0.357	&0.401	&0.421	&0.481	&	&0.037	&0.068	&0.090	&0.144	&0.229	&0.372	&0.295	&0.472
          \\
                        & Ext-L           & 0.005	&0.006	&0.169	&0.244	&0.398	&0.454	&0.421	&0.481	&	&0.045	&0.059	&0.083	&0.146	&0.196	&0.316	&0.295	&0.472
          \\
                        & Ext-V           & 0.004	&0.006	&0.246	&0.317	&0.398	&0.461	&0.421	&0.481	&	&0.049	&0.069	&0.105	&0.149	&0.224	&0.369	&0.295	&0.472
          \\
                        & BKD             & 0.075	&0.156	&0.343	&0.399	&0.414	&0.468
          & -              & -              &  & 0.113	&0.204	&0.244	&0.412	&0.287	&0.463
          & -              & -              \\
                        & TA              & 0.089	&0.189	&0.368	&0.418	&0.415	&0.472	&-	&-	&	&0.124	&0.221	&0.254	&0.416	&0.290	&0.465	&-	&-
              \\
                        & DualDE          & 0.083	&0.175	&\textbf{0.386}	&0.423	&\textbf{0.419}	&0.475	&-	&-	&	&0.120	&0.213	&0.258	&\textbf{0.429}	&0.293	&0.466	&-	&-
              \\
                        & IterDE  
                        &0.077	&0.162	&0.375	&0.419	&0.416	&0.469	&-	&-	&	&0.120	&0.215	&0.257	&0.427	&\textbf{0.293}	&0.465	&-	&-
\\
                        & {\model}        & \textbf{0.111}	&\textbf{0.224}	&0.385	&\textbf{0.431}	&0.418	
                        &\textbf{0.477}	&\textbf{0.421}
                        &\textbf{0.482}	&	&\textbf{0.143}	&\textbf{0.267}	&\textbf{0.261}	&0.427	&0.291	&\textbf{0.466}	&0.294	&0.470          \\
                        \midrule
\multirow{9}{*}{RotatE} & DT              & 0.172          & 0.418          & 0.456          & 0.556          & 0.471          & 0.567          & 0.476          & 0.575          &  & 0.254          & 0.424          & 0.312          & 0.495          & 0.322          & 0.506          & \textbf{0.325} & \textbf{0.515} \\
                        & Ext             & 0.299          & 0.378          & 0.437          & 0.516          & 0.467          & 0.549          & 0.476          & \textbf{0.575} &  & 0.138          & 0.245          & 0.251          & 0.410          & 0.291          & 0.465          & 0.325          & 0.515          \\
                        & Ext-L           & 0.206          & 0.277          & 0.399          & 0.487          & 0.445          & 0.541          & 0.476          & 0.575          &  & 0.135          & 0.243          & 0.221          & 0.365          & 0.280          & 0.453          & 0.325          & 0.515          \\
                        & Ext-V           & 0.261          & 0.377          & 0.337          & 0.471          & 0.416          & 0.532          & 0.476          & 0.575          &  & 0.160          & 0.281          & 0.238          & 0.393          & 0.288          & 0.458          & 0.325          & 0.515          \\
                        & BKD             & 0.175          & 0.434          & 0.457          & 0.556          & 0.472          & 0.570          & -              & -              &  & 0.277          & 0.442          & 0.314          & 0.503          & 0.322          & 0.510          & -              & -              \\
                        & TA              & 0.177          & 0.438          & 0.459          & 0.558          & 0.473          & 0.572          & -              & -              &  & 0.280          & 0.447          & 0.313          & 0.501          & 0.323          & 0.510          & -              & -              \\
                        & DualDE          & 0.179          & 0.440          & 0.462          & 0.559          & \textbf{0.473} & 0.573          & -              & -              &  & 0.282          & 0.449          & 0.315          & 0.502          & 0.322          & \textbf{0.512} & -              & -              \\
                        & IterDE  &0.176	&0.436	&0.459	&0.560	&0.471	&0.569	&-	&-	&	&0.276	&0.445	&0.317	&0.504	&0.323	&0.512	&-	&- \\
                        & {\model}        & \textbf{0.324} & \textbf{0.469} & \textbf{0.466} & \textbf{0.561} & 0.471          & \textbf{0.574} & \textbf{0.476} & 0.574          &  & \textbf{0.288} & \textbf{0.459} & \textbf{0.318} & \textbf{0.504} & \textbf{0.323} & 0.510          & 0.324          & 0.514          \\
                        \midrule
\multirow{9}{*}{PairRE} & DT              & 0.220          & 0.321          & 0.415          & 0.472          & 0.449          & 0.534          & \textbf{0.453} & \textbf{0.544} &  & 0.182          & 0.314          & 0.284          & 0.452          & 0.319          & 0.505          & \textbf{0.332} & \textbf{0.522} \\
                        & Ext             & 0.152          & 0.209          & 0.334          & 0.463          & 0.419          & 0.526          & 0.453          & 0.544          &  & 0.148          & 0.222          & 0.217          & 0.353          & 0.294          & 0.469          & 0.332          & 0.522          \\
                        & Ext-L           & 0.162          & 0.220          & 0.363          & 0.442          & 0.437          & 0.523          & 0.453          & 0.544          &  & 0.150          & 0.249          & 0.219          & 0.333          & 0.309          & 0.489          & 0.332          & 0.522          \\
                        & Ext-V           & 0.172          & 0.260          & 0.389          & 0.456          & 0.441          & 0.529          & 0.453          & 0.544          &  & 0.176          & 0.277          & 0.229          & 0.374          & 0.311          & 0.490          & 0.332          & 0.522          \\
                        & BKD             & 0.228          & 0.336          & 0.421          & 0.483          & 0.451          & 0.536          & -              & -              &  & 0.198          & 0.332          & 0.288          & 0.453          & 0.321          & 0.508          & -              & -              \\
                        & TA              & 0.245          & 0.340          & 0.426          & 0.487          & 0.452          & 0.537          & -              & -              &  & 0.208          & 0.346          & 0.292          & 0.455          & 0.323          & 0.509          & -              & -              \\
                        & DualDE          & 0.242          & 0.336          & 0.428          & 0.495          & \textbf{0.453} & 0.540          & -              & -              &  & 0.207          & 0.342          & 0.293          & 0.456          & \textbf{0.326} & \textbf{0.512} & -              & -              \\
                        & IterDE  &0.235	&0.336	&0.426	&0.495	&0.450	&0.538	&-	&-  &	&0.205	&0.340	&0.293	&0.462	&0.324	&0.508	&-	&- \\
                        & {\model}        & \textbf{0.317} & \textbf{0.376} & \textbf{0.433} & \textbf{0.502} & 0.451          & \textbf{0.541} & 0.451          & 0.542          &  & \textbf{0.239} & \textbf{0.384} & \textbf{0.303} & \textbf{0.466} & 0.324          & 0.510          & 0.330          & 0.520         \\ \bottomrule
\end{tabular}
}
\end{center}
\caption{MRR and Hit@10 (H10) of some dimensions on WN18RR (WN) and FB15K237 (FB).}
\label{table_main_wn_fb}
\end{table*}

\begin{table*}[htbp]
\setlength\tabcolsep{3pt}
\renewcommand\arraystretch{0.9}
\setlength{\abovecaptionskip}{-0.cm}
\setlength{\belowcaptionskip}{-0.1cm}
\begin{center}
    \resizebox{0.9\textwidth}{!}{
\begin{tabular}{lccccclccccclccccclccccc}
\toprule
            & \multicolumn{5}{c}{FB15k-237}                                                      &  & \multicolumn{5}{c}{WN18RR}                                                         &  & \multicolumn{5}{c}{CoDEx-L}                                                        &  & \multicolumn{5}{c}{YAGO3-10}                                                       \\ \cline{2-6} \cline{8-12} \cline{14-18} \cline{20-24} 
            & \textit{Dim} & \textit{\#P(M)} & \textit{MRR}   & \textit{Hit@10} & \textit{Effi}  &  & \textit{Dim} & \textit{\#P(M)} & \textit{MRR}   & \textit{Hit@10} & \textit{Effi}  &  & \textit{Dim} & \textit{\#P(M)} & \textit{MRR}   & \textit{Hit@10} & \textit{Effi}  &  & \textit{Dim} & \textit{\#P(M)} & \textit{MRR}   & \textit{Hit@10} & \textit{Effi}  \\
            \midrule
RotatE      & 1000         & 29.3            & 0.336          & 0.532           & 0.011          &  & 500          & 40.6            & 0.508          & 0.612           & 0.013          &  & 500          & 78              & 0.258          & 0.387           & 0.003          &  & 500          & 123.2           & 0.495          & 0.670           & 0.004          \\
\hline
RotatE      & 100          & 2.9             & 0.296          & 0.473           & 0.102          &  & 50           & 4.1             & 0.411          & 0.429           & 0.100          &  & 25           & 3.8             & 0.196          & 0.322           & 0.052          &  & 20           & 4.8             & 0.121          & 0.262           & 0.025          \\
+ NodePiece & 100          & 3.2             & 0.256          & 0.420           & 0.080          &  & 100          & 4.4             & 0.403          & 0.515           & 0.092          &  & 100          & 3.6             & 0.190          & 0.313           & 0.053          &  & 100          & 4.1             & 0.247          & 0.488           & 0.060          \\
+ EARL      & 150          & 1.8             & 0.310          & 0.501           & 0.172          &  & 200          & 3.8             & 0.440          & 0.527           & 0.116          &  & 100          & 2.1             & 0.238          & \textbf{0.390}  & \textbf{0.113} &  & 100          & 3               & 0.302          & 0.498           & \textbf{0.101} \\
+ \model    & 40           & 1.2             & \textbf{0.318} & \textbf{0.504}  & \textbf{0.265} &  & 40           & 3.2             & \textbf{0.466} & \textbf{0.561}  & \textbf{0.146} &  & 20           & 3.1             & \textbf{0.243} & 0.385           & 0.078          &  & 20           & 4.9             & \textbf{0.313} & \textbf{0.528}  & 0.064         \\ \bottomrule
\end{tabular}
}
\end{center}
\caption{Link prediction results on WN18RR, FB15K237, CoDEx-L and YAGO3-10.}
\label{table_lp_all}
\end{table*}
{\model} outperforms baselines in most settings, especially at extremely low dimensions. On WN18RR with $d$=10, {\model} achieves 
an improvement of $\textbf{14.9\%}$ and $\textbf{15.1\%}$ on TransE, $\textbf{8.4\%}$ and $\textbf{6.6\%}$ on RotatE, $\textbf{29.4\%}$ and $\textbf{10.6\%}$ on PairRE than the best MRR and Hit@10 of baselines. A similar trend is seen on FB15K237.
This success stems from {\model}'s rich knowledge sources for low-dimensional models: For sub-model $M_i$, $M_{i+1}$ is a direct teacher, and $M_{i+2}$ can indirectly influence $M_i$ via $M_{i+1}$. In theory, all higher-dimensional sub-models can transfer knowledge to lower-dimensional ones through stepwise propagation. 
Although this propagation may harm high-dimensional models with low-quality knowledge from low-dimensional ones, {\model}'s evolutionary improvement mechanism mitigates the damage, allowing high-dimensional models to compete with directly trained KGEs (Fig.~\ref{fig:all_dim_pairre}).
We also note that Ext-based methods are highly unstable, performing worse than DT in most cases, suggesting that dimension importance alone doesn't ensure sub-model performance. 
\begin{figure}[ht]
\vspace{-2mm}
 \setlength{\abovecaptionskip}{0.5mm}
 \setlength{\belowcaptionskip}{-0.7mm}
  {\includegraphics[width=0.48\linewidth]{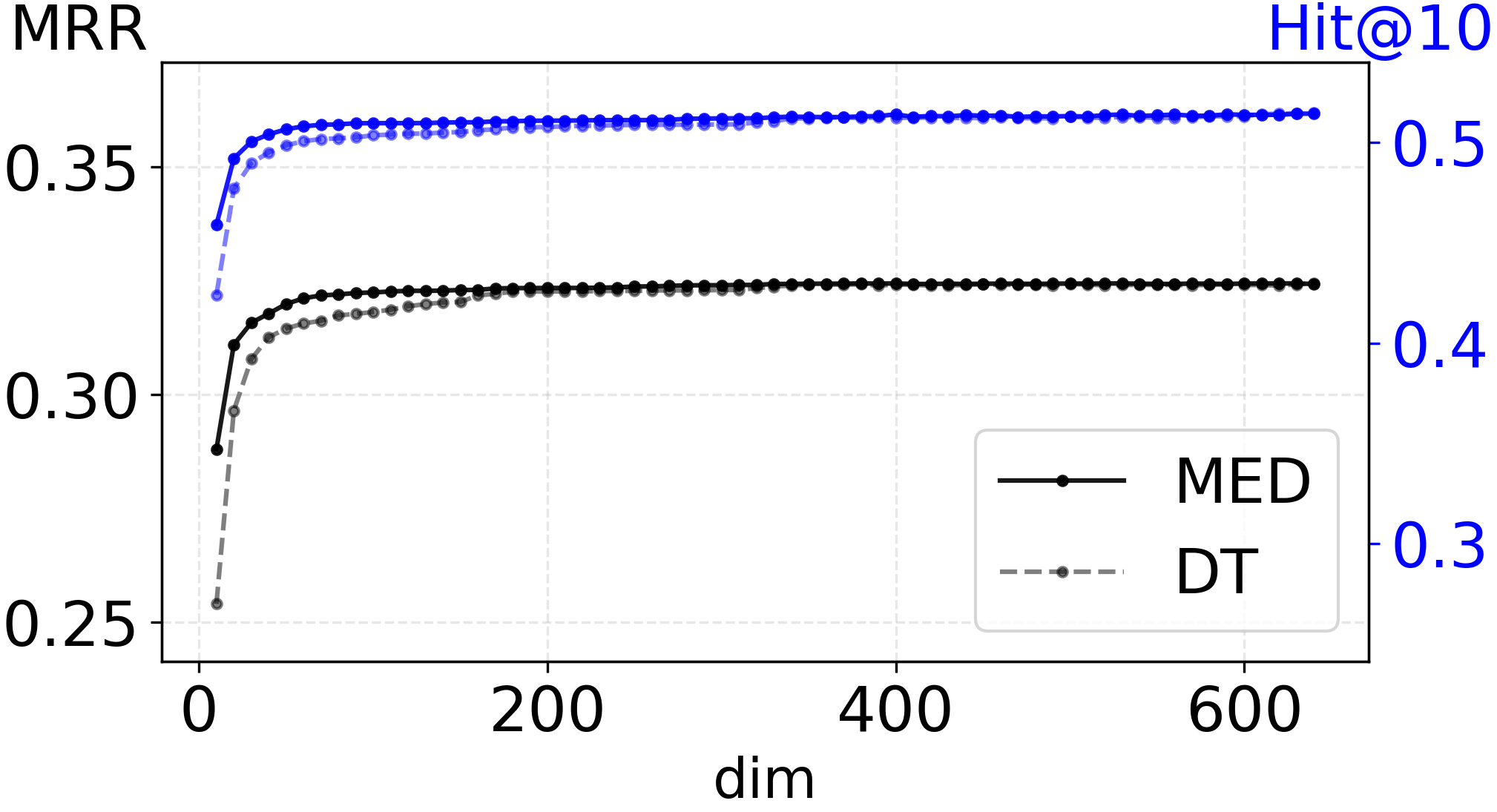}}
  {\includegraphics[width=0.48\linewidth]{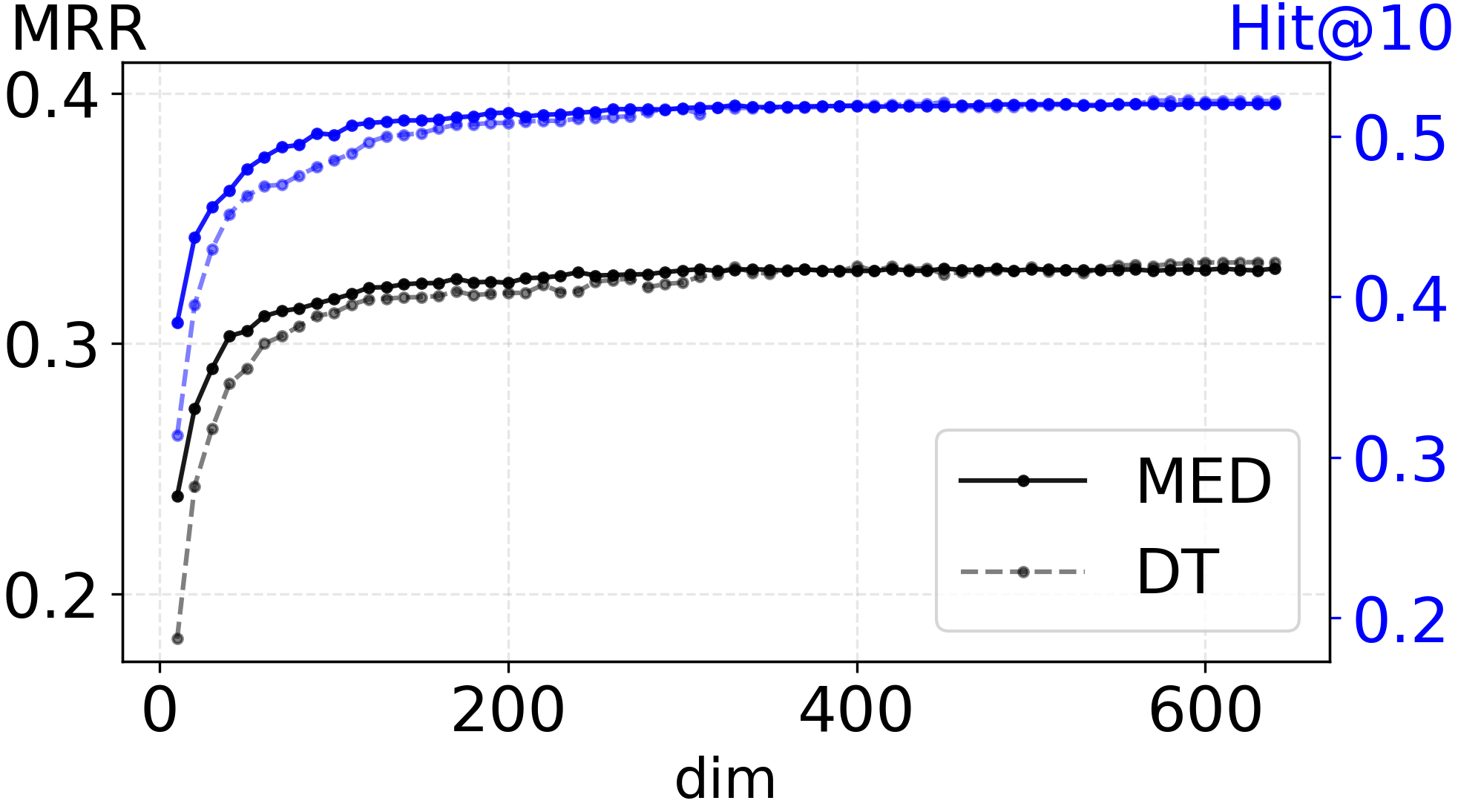}}
\caption{Results of different dimensions for PairRE on WN18RR (left) and FB15K237 (right).}
    \label{fig:all_dim_pairre}
\end{figure}
\subsection{Parameter efficiency of \model}
\label{sec:para-effi}
In Table~\ref{table_lp_all}, we compare our sub-models of suitable low dimensions to parameter-efficient KGEs especially proposed for large-scale KGs including NodePiece~\cite{DBLP:conf/iclr/0001DWH22} and EARL~\cite{DBLP:conf/aaai/ChenZYZGPC23}. 
With a similar number of model parameters, {\model}'s sub-models outperform specialized parameter-efficient KGE methods, showing their parameter efficiency. More crucially, it can offer different-sized parameter-efficient models for applications.
\begin{table}
\setlength{\abovecaptionskip}{0.0cm}    \setlength{\belowcaptionskip}{0.0cm}
    \centering
    \resizebox{0.45\textwidth}{!}{
    \begin{tabular}{lc|cc|cc|cc}
    \toprule  & 
     & \multicolumn{2}{c|}{\textbf{User Labeling}}  & \multicolumn{4}{c}{\textbf{Product Recommendation}}                                \\  & & \multicolumn{2}{c|}{server (500d)} & \multicolumn{2}{c|}{PC terminal (100d)} & \multicolumn{2}{c}{MP terminal (10d)} \\
     \midrule
     \textit{Method}    & \textit{train time} &\textit{acc.}     & \textit{f1}   & \textit{ndcg@5}   & \textit{ndcg@10}   & \textit{ndcg@5}     & \textit{ndcg@10}    \\
     DT                 & 103h & 0.889                 & 0.874               & 0.411             & 0.441              & 0.344               & 0.361               \\
PCA     &   -         & -                     & -                   & 0.417             & 0.447              & 0.392               & 0.418               \\
DualDE     &195h        & -                     & -                   & 0.423             & 0.456              & 0.404               & 0.433               \\
{\model}  &    53h     & \textbf{0.893}        & \textbf{0.879}      & \textbf{0.431}    & \textbf{0.465}     & \textbf{0.422}      & \textbf{0.451}     \\
\bottomrule
\vspace{-2mm}
\end{tabular}
    }
    \caption{Results on SKG.}
    \label{tab:downstream}
\end{table}
\subsection{{\model} in real applications}
\label{sec:application}
We apply the trained croppable KGE with TransE on SKG to three real-world applications: user labeling on servers and product recommendation on PCs and mobile phones. Table~\ref{tab:downstream} shows that our croppable user embeddings outperform all baselines, including directly trained (DT), the best baseline DualDE, and a common dimension reduction method in industry principal components analysis (PCA) on $M^{DT}_{max}$. 
Notably, the strong performance in the mobile phone task (constrained by storage and computing resources to a max embedding dimension of 10) highlights the great practical value of MED. More application details are in Appendix~\ref{sec:detail_application}.

\begin{table*}[h]
\setlength\tabcolsep{4pt}
\renewcommand\arraystretch{0.9}
\setlength{\abovecaptionskip}{-0.1cm}
\setlength{\belowcaptionskip}{-0.0cm}
\begin{center}
        \resizebox{.7\textwidth}{!}{
\begin{tabular}{lcccccccccc}
\toprule
\textit{Method}            & \textit{\#P(M)} & \textit{Speedup}                   & \begin{tabular}[c]{@{}c@{}}MNLI-m\\ \textit{acc.}\end{tabular} & \begin{tabular}[c]{@{}c@{}}MNLI-mm\\ \textit{acc.}\end{tabular} & \begin{tabular}[c]{@{}c@{}}MRPC\\ \textit{f1/acc.}\end{tabular} & \begin{tabular}[c]{@{}c@{}}QNLI\\ \textit{acc.}\end{tabular} & \begin{tabular}[c]{@{}c@{}}QQP\\ \textit{f1/acc.}\end{tabular} & \begin{tabular}[c]{@{}c@{}}RTE\\ \textit{acc.}\end{tabular} & \begin{tabular}[c]{@{}c@{}}STS-2\\ \textit{acc.}\end{tabular} & \begin{tabular}[c]{@{}c@{}}STS-B\\ \textit{pear./spear.}\end{tabular}\\ \midrule
BERT$_{Base}^{\dagger}$         & 110    & 1.0$\times$                                                    & 84.4                                                    & 85.3                                                     & 88.6/84.1                                               & 89.7                                                  & 89.6/91.1                                              & 67.5                                                 & 92.5 & 88.8/88.5                                                  \\ \midrule
BERT$_{6}$\textnormal{-BKD}        & 66     & 2.0$\times$                                                    & 82.2                                                    & 82.9                                                     & 86.2/80.8                                               & 88.5                                                  & 88.0/91.0                                              & 65.4                                                 & 90.9  &88.2/87.8                                                  \\
BERT$_{6}$\textnormal{-PKD}        & 66     & 2.0$\times$                                                    & 82.3                                                    & 82.6                                                     & 86.4/81.0                                               & 88.6                                                  & 87.9/91.0                                              & 63.9                                                 & 90.8  &88.5/88.1                                                 \\
BERT$_{6}$\textnormal{-MiniLM}     & 66     & 2.0$\times$                                                    & 82.2                                                    & 82.6                                                     & 84.6/78.1                                               & \textbf{89.5}                                                  & 87.2/90.5                                              & 61.5                                                 & 90.2  &87.8/87.5                                                 \\
BERT$_{6}$\textnormal{-RKD}        & 66     & 2.0$\times$                                                    & 82.4                                                    & 82.9                                                     & 86.9/81.8                                               & 88.9                                                  & 88.1/\textbf{91.2}                                              & 65.2                                                 & 91.0   &88.4/88.1                                                \\
BERT$_{6}$\textnormal{-FSD}        & 66     & 2.0$\times$                                                    & 82.4                                                    & 83.0                                                     & 87.1/82.2                                               & 89.0                                                  & 88.1/\textbf{91.2}                                              & 66.6                                                 & 91.0    &\textbf{88.7/88.3}                                               \\
BERT$_{4}$\textnormal{-BKD}          & 55     & 2.9$\times$                                                    & 80.5                                                    & 80.9                                                     & 87.2/83.1                                               & 87.5                                                  & 86.6/90.4                                              & 65.2                                                 & 90.2    &84.5/84.2                                               \\
BERT$_{4}$\textnormal{-PKD}         & 55     & 2.9$\times$                                                    & 80.9                                                    & 81.3                                                     & 87.0/82.9                                               & 87.7                                                  & 86.8/90.5                                              & 66.1                                                 & 90.5     &84.3/84.0                                                \\
BERT$_{4}$\textnormal{-MetaDistil} & 55     & 2.9$\times$                                                    & 82.4                                                    & 82.7                                                     & \textbf{88.4/84.2}                                               & 88.6                                                  & 87.8/90.8                                              & \textbf{67.8}                                                 & 91.8  &86.3/86.0                                                 \\
BERT-HAT$^{\dagger}$       & 54     & 2.0$\times$                                                        & 70.8                                                    & 71.6                                                     & 81.2/74.8                                                   & 65.3                                                  & 76.1/80.4                                                   & 52.7                                                 & 84.3                   & 79.6/80.1                                \\
BERT-MED   & 54     & 2.0$\times$                                                       & \textbf{82.7}                                                    & \textbf{83.3}                                                     & 88.0/84.0                                                   & 86.8                                                  & \textbf{89.1}/90.7                                              & 67.2                                                 & \textbf{91.9}      & 87.6/87.2                                             \\ \midrule
BERT-HAT$^{\dagger}$          & 17.5   & 4.7$\times$                                                                      & 63.6                                                    & 64.2                                                     & 68.4/78.4                                                    & 61.1                                                  & 69.0/79.7                                                   & 47.2                                                 & 82.9      &74.1/75.8                                             \\
BERT-MED   & 17.5   & 4.7$\times$                                                                           & 81.2                                                    & 82.4                                                     & 86.1/82.0                                                    & 86.4                                                  & 83.8/86.2                                              & 64.6                                                 & 88.2    &86.1/86.4                                              \\ \midrule
BERT-HAT$^{\dagger}$         & 6.36   & 5.2$\times$                                                                       & 59.9                                                    & 60.0                                                     & 66.5/77.3                                                    & 60.1                                                  & 66.5/77.1                                                   & 46.2                                                 & 81.7             & 71.9/70.4                                      \\
BERT-MED   & 6.36   & 5.2$\times$                                                                         & 72.6                                                    & 73.7                                                     & 84.1/78.1                                                    & 86.0                                                  & 79.6/82.7                                              & 61.7                                                 & 86.9     & 82.8/81.6                                      \\ \bottomrule

\end{tabular}
}
\end{center}
\caption{Results on the dev set of GLUE. The results of knowledge distillation methods for BERT$_{4}$ and BERT$_{6}$ are reported by \cite{DBLP:journals/eswa/JungKNK23,DBLP:conf/acl/ZhouXM22} and the $^{\dagger}$results reported by us.}
\label{table_glue}
\vspace{-0.35cm}
\end{table*}
\subsection{Extend {\model} to Neural Networks}
\label{sec:extend_to_bert}
To test our method's extensibility to other neural networks, we use BERT~\cite{DBLP:conf/naacl/DevlinCLT19} as an example. 
We use distillation methods based on Hugging Face Transformers \cite{DBLP:conf/emnlp/WolfDSCDMCRLFDS20} as baselines. As in previous works\cite{DBLP:conf/emnlp/SunCGL19,DBLP:journals/corr/abs-1903-12136,DBLP:journals/eswa/JungKNK23,DBLP:conf/acl/ZhouXM22}, we perform distillation at the fine-tuning stage. See Appendix~\ref{sec:detail_bert} for more details.
Table~\ref{table_glue} shows the results on the GLUE development set\cite{DBLP:conf/iclr/WangSMHLB19}. 
We compare {\model} with other KD models of similar speedup or parameter number. MED performs competitively on most tasks against BERT-specialized KD methods. 
Compared to HAT \cite{DBLP:conf/acl/WangWLCZGH20}, which has a similar architecture to ours, {\model}'s sub-models outperform HAT across three parameter levels. Sub-models with 54M, 17.5M, and 6.36M parameters show average improvements of $16.3\%$, $21.7\%$ and $19.7\%$ respectively.

\subsection{Analysis of {\model}}
\label{sec:analysis}
\subsubsection{Training efficiency}
\label{sec:exper_time}
\begin{table}
\setlength\tabcolsep{3pt}
\setlength{\abovecaptionskip}{-0.cm}    \setlength{\belowcaptionskip}{-0.2cm}
\begin{center}
    \resizebox{0.48\textwidth}{!}{
\begin{tabular}{ll|rr|rr|rr|rr}
\toprule
                    &           & \multicolumn{2}{c}{\textbf{TransE}} & \multicolumn{2}{c}{\textbf{SimplE}} & \multicolumn{2}{c}{\textbf{RotatE}} & \multicolumn{2}{c}{\textbf{PairRE}} \\
                    \midrule
\multirow{7}{*}{WN} & DT        & 74.0        & (9.49$\times$)        & 68.0        & (12.14$\times$)       & 141.0       & (11.10$\times$)       & 67.4        & (10.06$\times$)       \\
                    & Ext-based & 1.5         & (0.19$\times$)        & 1.3         & (0.23$\times$)        & 2.5         & (0.20$\times$)        & 1.6         & (0.24$\times$)        \\
                    & BKD       & 91.5        & (11.73$\times$)       & 72.0        & (12.86$\times$)       & 163.0       & (12.83$\times$)       & 87.5        & (13.06$\times$)       \\
                    & TA        & 172.0       & (22.05$\times$)       & 142.0       & (25.36$\times$)       & 272.0       & (21.42$\times$)       & 166.0       & (24.78$\times$)       \\
                    & DualDE    & 151.0       & (19.36$\times$)       & 133.0       & (23.75$\times$)       & 240.0       & (18.90$\times$)       & 133.0       & (19.85$\times$)       \\
                    & IterDE    & 140.9       & (18.06$\times$)       & 118.0       & (21.07$\times$)       & 216.0       & (17.01$\times$)       & 124.0       & (18.51$\times$)       \\
                    & {\model}  & 7.8         & (1.00$\times$)        & 5.6         & (1.00$\times$)        & 12.7        & (1.00$\times$)        & 6.7         & (1.00$\times$)        \\
                    \midrule
\multirow{4}{*}{FB} & DT        & 218.0       & (10.23$\times$)       & 179.0       & (10.65$\times$)       & 381.0       & (10.73$\times$)       & 179.0       & (9.37$\times$)        \\
                    & Ext-based & 4.7         & (0.22$\times$)        & 5.1         & (0.30$\times$)        & 9.5         & (0.27$\times$)        & 3.7         & (0.19$\times$)        \\
                    & BKD       & 248.0       & (11.64$\times$)       & 227.0       & (13.51$\times$)       & 443.0       & (12.48$\times$)       & 231.0       & (12.09$\times$)       \\
                    & {\model}  & 21.3        & (1.00$\times$)        & 16.8        & (1.00$\times$)        & 35.5        & (1.00$\times$)        & 19.1        & (1.00$\times$)       \\
                    \bottomrule
\end{tabular}
}
\end{center}
\caption{Training time (hours).}\label{table_efficiency}
\end{table}
\begin{figure}[ht]
\centering
\vspace{-2mm}
 \setlength{\abovecaptionskip}{-0.cm}
 \setlength{\belowcaptionskip}{-0.3cm}
  {\includegraphics[width=0.5\linewidth]{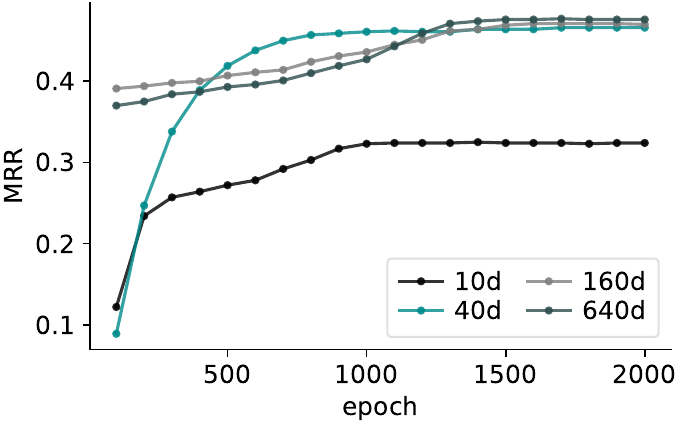}}
\caption{Sub-models' MRR during training on WN18RR with RotatE.}
    \label{fig:mrr_epoch}
    \vspace{-0cm}
\end{figure}
Table~\ref{table_efficiency} reports the training time of getting 64 models of various sizes ($d$=10, 20, ..., 640). Figure~\ref{fig:mrr_epoch} shows MRR change of MED's sub-models during training.
For fair comparison, all training is done on a single NVIDIA Tesla A100 40GB GPU. For DT, the training time is the sum of directly training 64 KGE models sequentially. Ext-based baselines have the same training time, equal to that of training a $d_{n}$-dimensional KGE model as dimension arrangement time is negligible. KD-based baselines' training time is the sum of training a $d_{n}$-dimensional teacher model and distilling 63 student models. 
On FB15K237, we don't train all 63 sizes of student models for TA, DualDE, and IterDE, estimated to take over 400 hours per KGE method.
Compared to DT, {\model} speeds up by up to 10$\times$ for 4 KGE methods. Ext-based baselines have the shortest training time but poor performance and low practical value. Except for BKD, other KD-based methods need to optimize both student and teacher, increasing train cost.

\subsubsection{Effect of the number of sub-models}
\begin{table}
\setlength\tabcolsep{3pt}
\setlength{\abovecaptionskip}{-0.cm}
\setlength{\belowcaptionskip}{-0.cm}
\begin{center}
\resizebox{0.45\textwidth}{!}{
\begin{tabular}{c|c|cc|cc|cc|cc}
\toprule
           &                        & \multicolumn{2}{c}{\textbf{10d}} & \multicolumn{2}{c}{\textbf{40d}} & \multicolumn{2}{c}{\textbf{160d}} & \multicolumn{2}{c}{\textbf{640d}} \\
\textit{n} & \textit{train time} & \textit{MRR}  & \textit{H10}  & \textit{MRR}  & \textit{H10}  & \textit{MRR}   & \textit{H10}  & \textit{MRR}   & \textit{H10}  \\
\midrule
64         & 12.7h                  & 0.324         & 0.469            & 0.466         & 0.561            & 0.471          & 0.574            & 0.476          & 0.574            \\
16         & 6.2h                   & 0.322         & 0.467            & 0.465         & 0.561            & 0.473          & 0.575            & 0.477          & 0.576            \\
4          & 3.3h                   & 0.319         & 0.463            & 0.463         & 0.561            & 0.475          & 0.577            & 0.480          & 0.578           \\
\bottomrule
\end{tabular}
}
\end{center}
\caption{Results of different $n$.}
\label{table_different_n}
\vspace{-0.3cm}
\end{table}
We set the number of different sub-models ($n$= 64, 16, 4) for RotatE on WN18RR. Table~\ref{table_different_n} shows that reducing the number of sub - models boosts high-dimensional ($d$=160 and 640) model performance but lowers that of low-dimensional ($d$=10 and 40) models.
Training efficiency is nearly linearly related to the number of models.

\subsubsection{Whether high-dimensional sub-models cover low-dimensional ones' capabilities}
A high-dimensional model retaining lower-dimensional ones' ability should correctly predict all triples the latter can. 
We calculate the percentage of test-set triples where if the smallest sub-model correctly predicting a triple is $M_i$, all higher-dimensional ones ($M_{i+1}$, $M_{i+2}$, ..., $M_n$) do too. 
This result is the ability retention ratio (ARR). 
We use Hit@10 to judge correct prediction: $M_i$ predicts a triple correctly if it scores in the top 10 among candidates.
\begin{figure}
\centering  
\setlength{\abovecaptionskip}{-0.0cm}
\setlength{\belowcaptionskip}{-0.2cm}
\includegraphics[width=1\linewidth]{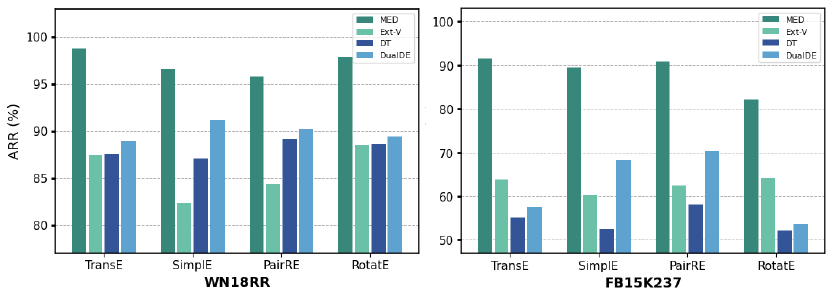}
\caption{The ability retention ratio (ARR).}
\vspace{-0.cm}
\label{fig:overlap}
\end{figure}
Fig.~\ref{fig:overlap} shows that {\model}'s ARR is is much higher than baselines, especially on FB15K237. This means {\model}'s high-dimensional sub-models cover low-dimensional ones' power, thanks to the mutual learning mechanism for knowledge review.
This advantage of {\model} provides a simple way to judge triple difficulty for KGE methods: triples low-dimensional sub-models predict may be easy as high-dimensional ones can too, while those only predicted by high-dimensional sub-models are difficult.

\section{Conclusion}
In this work, we propose a novel KGE training framework, {\model}, which trains a croppable KGE in one go. Sub-models of various dimensions can be cropped from it and used directly without extra training. 
In {\model}, we introduce the mutual learning mechanism to improve low-dimensional sub-models and make the high-dimensional sub-models retain low-dimensional ones' ability, the evolutionary improvement mechanism to prompt high-dimensional sub-models to learn what low-dimensional ones can't, and dynamic loss weights to balance multiple losses adaptively. 
Experimental results demonstrate MED's effectiveness, high efficiency, and flexible extensibility. 

\section{Limitations}
As a research paper, we are unable to conduct experiments on all knowledge graph embedding (KGE) methods. To demonstrate the generality of the method proposed in this paper, we selected representative models from different types of KGE methods, including distance-based TransE, rotation-based RotatE, semantic matching-based SimplE, and subrelation encoding-based PairRE. In fact, there are many KGE methods that outperform those used in our experiments. However, the method proposed in this paper is applicable to any KGE method with a triple scoring function.

\section{Acknowledgements}
This work is founded by National Natural Science Foundation of China (NSFCU23B2055/NSFCU19B2027/NSFC62306276), Zhejiang Provincial Natural Science Foundation of China (No. LQ23F020017), Yongjiang Talent Introduction Programme (2022A-238-G), and Fundamental Research Funds for the Central Universities (226-2023-00138). This work was supported by AntGroup.

\bibliography{custom}

\begin{thebibliography}{59}
\expandafter\ifx\csname natexlab\endcsname\relax\def\natexlab#1{#1}\fi

\bibitem[{Bai et~al.(2021)Bai, Zhang, Hou, Shang, Jin, Jiang, Liu, Lyu, and King}]{DBLP:conf/acl/BaiZHSJJLLK20}
Haoli Bai, Wei Zhang, Lu~Hou, Lifeng Shang, Jin Jin, Xin Jiang, Qun Liu, Michael~R. Lyu, and Irwin King. 2021.
\newblock Binarybert: Pushing the limit of {BERT} quantization.
\newblock In \emph{{ACL/IJCNLP} {(1)}}, pages 4334--4348. Association for Computational Linguistics.

\bibitem[{Bordes et~al.(2013)Bordes, Usunier, Garc{\'{\i}}a{-}Dur{\'{a}}n, Weston, and Yakhnenko}]{DBLP:conf/nips/BordesUGWY13}
Antoine Bordes, Nicolas Usunier, Alberto Garc{\'{\i}}a{-}Dur{\'{a}}n, Jason Weston, and Oksana Yakhnenko. 2013.
\newblock Translating embeddings for modeling multi-relational data.
\newblock In \emph{{NIPS}}, pages 2787--2795.

\bibitem[{Chao et~al.(2021)Chao, He, Wang, and Chu}]{DBLP:conf/acl/ChaoHWC20}
Linlin Chao, Jianshan He, Taifeng Wang, and Wei Chu. 2021.
\newblock Pairre: Knowledge graph embeddings via paired relation vectors.
\newblock In \emph{{ACL/IJCNLP} {(1)}}, pages 4360--4369. Association for Computational Linguistics.

\bibitem[{Chen et~al.(2023)Chen, Zhang, Yao, Zhu, Gao, Pan, and Chen}]{DBLP:conf/aaai/ChenZYZGPC23}
Mingyang Chen, Wen Zhang, Zhen Yao, Yushan Zhu, Yang Gao, Jeff~Z. Pan, and Huajun Chen. 2023.
\newblock Entity-agnostic representation learning for parameter-efficient knowledge graph embedding.
\newblock In \emph{{AAAI}}, pages 4182--4190. {AAAI} Press.

\bibitem[{Conneau and Kiela(2018)}]{DBLP:conf/lrec/ConneauK18}
Alexis Conneau and Douwe Kiela. 2018.
\newblock \href {http://www.lrec-conf.org/proceedings/lrec2018/summaries/757.html} {Senteval: An evaluation toolkit for universal sentence representations}.
\newblock In \emph{Proceedings of the Eleventh International Conference on Language Resources and Evaluation, {LREC} 2018, Miyazaki, Japan, May 7-12, 2018}. European Language Resources Association {(ELRA)}.

\bibitem[{Daiber et~al.(2013)Daiber, Jakob, Hokamp, and Mendes}]{DBLP:conf/i-semantics/DaiberJHM13}
Joachim Daiber, Max Jakob, Chris Hokamp, and Pablo~N. Mendes. 2013.
\newblock Improving efficiency and accuracy in multilingual entity extraction.
\newblock In \emph{{I-SEMANTICS}}, pages 121--124. {ACM}.

\bibitem[{Dettmers et~al.(2018)Dettmers, Minervini, Stenetorp, and Riedel}]{DBLP:conf/aaai/DettmersMS018}
Tim Dettmers, Pasquale Minervini, Pontus Stenetorp, and Sebastian Riedel. 2018.
\newblock Convolutional 2d knowledge graph embeddings.
\newblock In \emph{{AAAI}}, pages 1811--1818. {AAAI} Press.

\bibitem[{Devlin et~al.(2019)Devlin, Chang, Lee, and Toutanova}]{DBLP:conf/naacl/DevlinCLT19}
Jacob Devlin, Ming{-}Wei Chang, Kenton Lee, and Kristina Toutanova. 2019.
\newblock {BERT:} pre-training of deep bidirectional transformers for language understanding.
\newblock In \emph{{NAACL-HLT} {(1)}}, pages 4171--4186. Association for Computational Linguistics.

\bibitem[{Diefenbach et~al.(2018)Diefenbach, Singh, and Maret}]{DBLP:conf/www/DiefenbachSM18}
Dennis Diefenbach, Kamal~Deep Singh, and Pierre Maret. 2018.
\newblock Wdaqua-core1: {A} question answering service for {RDF} knowledge bases.
\newblock In \emph{{WWW} (Companion Volume)}, pages 1087--1091. {ACM}.

\bibitem[{Dolan and Brockett(2005)}]{DBLP:conf/acl-iwp/DolanB05}
William~B. Dolan and Chris Brockett. 2005.
\newblock \href {https://aclanthology.org/I05-5002/} {Automatically constructing a corpus of sentential paraphrases}.
\newblock In \emph{Proceedings of the Third International Workshop on Paraphrasing, IWP@IJCNLP 2005, Jeju Island, Korea, October 2005, 2005}. Asian Federation of Natural Language Processing.

\bibitem[{Fan et~al.(2024)Fan, Chen, Xue, Kong, Tao, and Lv}]{DBLP:conf/aaai/FanCXK0L24}
Cunhang Fan, Yujie Chen, Jun Xue, Yonghui Kong, Jianhua Tao, and Zhao Lv. 2024.
\newblock Progressive distillation based on masked generation feature method for knowledge graph completion.
\newblock In \emph{{AAAI}}, pages 8380--8388. {AAAI} Press.

\bibitem[{Galkin et~al.(2022)Galkin, Denis, Wu, and Hamilton}]{DBLP:conf/iclr/0001DWH22}
Mikhail Galkin, Etienne~G. Denis, Jiapeng Wu, and William~L. Hamilton. 2022.
\newblock Nodepiece: Compositional and parameter-efficient representations of large knowledge graphs.
\newblock In \emph{{ICLR}}. OpenReview.net.

\bibitem[{Han et~al.(2018)Han, Cao, Lv, Lin, Liu, Sun, and Li}]{DBLP:conf/emnlp/HanCLLLSL18}
Xu~Han, Shulin Cao, Xin Lv, Yankai Lin, Zhiyuan Liu, Maosong Sun, and Juanzi Li. 2018.
\newblock Openke: An open toolkit for knowledge embedding.
\newblock In \emph{{EMNLP} (Demonstration)}, pages 139--144. Association for Computational Linguistics.

\bibitem[{He et~al.(2017)He, Liao, Zhang, Nie, Hu, and Chua}]{DBLP:conf/www/HeLZNHC17}
Xiangnan He, Lizi Liao, Hanwang Zhang, Liqiang Nie, Xia Hu, and Tat{-}Seng Chua. 2017.
\newblock Neural collaborative filtering.
\newblock In \emph{{WWW}}, pages 173--182. {ACM}.

\bibitem[{Hinton et~al.(2015)Hinton, Vinyals, and Dean}]{DBLP:journals/corr/HintonVD15}
Geoffrey~E. Hinton, Oriol Vinyals, and Jeffrey Dean. 2015.
\newblock Distilling the knowledge in a neural network.
\newblock \emph{CoRR}, abs/1503.02531.

\bibitem[{Huber and Peter(1964)}]{Huber1964Robust}
Huber and J.~Peter. 1964.
\newblock Robust estimation of a location parameter.
\newblock \emph{The Annals of Mathematical Statistics}, 35(1):73--101.

\bibitem[{Ji et~al.(2015)Ji, He, Xu, Liu, and Zhao}]{DBLP:conf/acl/JiHXL015}
Guoliang Ji, Shizhu He, Liheng Xu, Kang Liu, and Jun Zhao. 2015.
\newblock Knowledge graph embedding via dynamic mapping matrix.
\newblock In \emph{{ACL} {(1)}}, pages 687--696. The Association for Computer Linguistics.

\bibitem[{Jung et~al.(2023)Jung, Kim, Na, and Kim}]{DBLP:journals/eswa/JungKNK23}
Hee{-}Jun Jung, Doyeon Kim, Seung{-}Hoon Na, and Kangil Kim. 2023.
\newblock \href {https://doi.org/10.1016/J.ESWA.2023.120980} {Feature structure distillation with centered kernel alignment in {BERT} transferring}.
\newblock \emph{Expert Syst. Appl.}, 234:120980.

\bibitem[{Kazemi and Poole(2018)}]{DBLP:conf/nips/Kazemi018}
Seyed~Mehran Kazemi and David Poole. 2018.
\newblock Simple embedding for link prediction in knowledge graphs.
\newblock In \emph{NeurIPS}, pages 4289--4300.

\bibitem[{Kingma and Ba(2015)}]{DBLP:journals/corr/KingmaB14}
Diederik~P. Kingma and Jimmy Ba. 2015.
\newblock Adam: {A} method for stochastic optimization.
\newblock In \emph{{ICLR} (Poster)}.

\bibitem[{Lin et~al.(2015)Lin, Liu, Sun, Liu, and Zhu}]{DBLP:conf/aaai/LinLSLZ15}
Yankai Lin, Zhiyuan Liu, Maosong Sun, Yang Liu, and Xuan Zhu. 2015.
\newblock Learning entity and relation embeddings for knowledge graph completion.
\newblock In \emph{{AAAI}}, pages 2181--2187. {AAAI} Press.

\bibitem[{Liu et~al.(2024)Liu, Ke, Wang, Shang, Gao, Li, Ji, and Liu}]{DBLP:conf/aaai/LiuK0SGLJL24}
Jiajun Liu, Wenjun Ke, Peng Wang, Ziyu Shang, Jinhua Gao, Guozheng Li, Ke~Ji, and Yanhe Liu. 2024.
\newblock Towards continual knowledge graph embedding via incremental distillation.
\newblock In \emph{{AAAI}}, pages 8759--8768. {AAAI} Press.

\bibitem[{Liu et~al.(2023)Liu, Wang, Shang, and Wu}]{DBLP:conf/aaai/LiuWSW23}
Jiajun Liu, Peng Wang, Ziyu Shang, and Chenxiao Wu. 2023.
\newblock Iterde: An iterative knowledge distillation framework for knowledge graph embeddings.
\newblock In \emph{{AAAI}}, pages 4488--4496. {AAAI} Press.

\bibitem[{Mahdisoltani et~al.(2015)Mahdisoltani, Biega, and Suchanek}]{DBLP:conf/cidr/MahdisoltaniBS15}
Farzaneh Mahdisoltani, Joanna Biega, and Fabian~M. Suchanek. 2015.
\newblock {YAGO3:} {A} knowledge base from multilingual wikipedias.
\newblock In \emph{{CIDR}}. www.cidrdb.org.

\bibitem[{Mirzadeh et~al.(2020)Mirzadeh, Farajtabar, Li, Levine, Matsukawa, and Ghasemzadeh}]{DBLP:conf/aaai/MirzadehFLLMG20}
Seyed{-}Iman Mirzadeh, Mehrdad Farajtabar, Ang Li, Nir Levine, Akihiro Matsukawa, and Hassan Ghasemzadeh. 2020.
\newblock Improved knowledge distillation via teacher assistant.
\newblock In \emph{{AAAI}}, pages 5191--5198. {AAAI} Press.

\bibitem[{Molchanov et~al.(2017)Molchanov, Tyree, Karras, Aila, and Kautz}]{DBLP:conf/iclr/MolchanovTKAK17}
Pavlo Molchanov, Stephen Tyree, Tero Karras, Timo Aila, and Jan Kautz. 2017.
\newblock Pruning convolutional neural networks for resource efficient inference.
\newblock In \emph{{ICLR} (Poster)}. OpenReview.net.

\bibitem[{Nguyen et~al.(2018)Nguyen, Nguyen, Nguyen, and Phung}]{DBLP:conf/naacl/NguyenNNP18}
Dai~Quoc Nguyen, Tu~Dinh Nguyen, Dat~Quoc Nguyen, and Dinh~Q. Phung. 2018.
\newblock A novel embedding model for knowledge base completion based on convolutional neural network.
\newblock In \emph{{NAACL-HLT} {(2)}}, pages 327--333. Association for Computational Linguistics.

\bibitem[{Nickel et~al.(2011)Nickel, Tresp, and Kriegel}]{DBLP:conf/icml/NickelTK11}
Maximilian Nickel, Volker Tresp, and Hans{-}Peter Kriegel. 2011.
\newblock A three-way model for collective learning on multi-relational data.
\newblock In \emph{{ICML}}, pages 809--816. Omnipress.

\bibitem[{Park et~al.(2019)Park, Kim, Lu, and Cho}]{DBLP:conf/cvpr/ParkKLC19}
Wonpyo Park, Dongju Kim, Yan Lu, and Minsu Cho. 2019.
\newblock Relational knowledge distillation.
\newblock In \emph{{CVPR}}, pages 3967--3976. Computer Vision Foundation / {IEEE}.

\bibitem[{Rajpurkar et~al.(2016)Rajpurkar, Zhang, Lopyrev, and Liang}]{DBLP:conf/emnlp/RajpurkarZLL16}
Pranav Rajpurkar, Jian Zhang, Konstantin Lopyrev, and Percy Liang. 2016.
\newblock \href {https://doi.org/10.18653/V1/D16-1264} {Squad: 100, 000+ questions for machine comprehension of text}.
\newblock In \emph{Proceedings of the 2016 Conference on Empirical Methods in Natural Language Processing, {EMNLP} 2016, Austin, Texas, USA, November 1-4, 2016}, pages 2383--2392. The Association for Computational Linguistics.

\bibitem[{Sachan(2020)}]{DBLP:conf/acl/Sachan20}
Mrinmaya Sachan. 2020.
\newblock Knowledge graph embedding compression.
\newblock In \emph{{ACL}}, pages 2681--2691. Association for Computational Linguistics.

\bibitem[{Safavi and Koutra(2020)}]{DBLP:conf/emnlp/SafaviK20}
Tara Safavi and Danai Koutra. 2020.
\newblock Codex: {A} comprehensive knowledge graph completion benchmark.
\newblock In \emph{{EMNLP} {(1)}}, pages 8328--8350. Association for Computational Linguistics.

\bibitem[{Schlichtkrull et~al.(2018)Schlichtkrull, Kipf, Bloem, van~den Berg, Titov, and Welling}]{DBLP:conf/esws/SchlichtkrullKB18}
Michael~Sejr Schlichtkrull, Thomas~N. Kipf, Peter Bloem, Rianne van~den Berg, Ivan Titov, and Max Welling. 2018.
\newblock Modeling relational data with graph convolutional networks.
\newblock In \emph{{ESWC}}, volume 10843 of \emph{Lecture Notes in Computer Science}, pages 593--607. Springer.

\bibitem[{Socher et~al.(2013)Socher, Perelygin, Wu, Chuang, Manning, Ng, and Potts}]{DBLP:conf/emnlp/SocherPWCMNP13}
Richard Socher, Alex Perelygin, Jean Wu, Jason Chuang, Christopher~D. Manning, Andrew~Y. Ng, and Christopher Potts. 2013.
\newblock \href {https://aclanthology.org/D13-1170/} {Recursive deep models for semantic compositionality over a sentiment treebank}.
\newblock In \emph{Proceedings of the 2013 Conference on Empirical Methods in Natural Language Processing, {EMNLP} 2013, 18-21 October 2013, Grand Hyatt Seattle, Seattle, Washington, USA, {A} meeting of SIGDAT, a Special Interest Group of the {ACL}}, pages 1631--1642. {ACL}.

\bibitem[{Stock et~al.(2021)Stock, Fan, Graham, Grave, Gribonval, J{\'{e}}gou, and Joulin}]{DBLP:conf/iclr/StockFGGGJJ21}
Pierre Stock, Angela Fan, Benjamin Graham, Edouard Grave, R{\'{e}}mi Gribonval, Herv{\'{e}} J{\'{e}}gou, and Armand Joulin. 2021.
\newblock Training with quantization noise for extreme model compression.
\newblock In \emph{{ICLR}}. OpenReview.net.

\bibitem[{Sun et~al.(2019{\natexlab{a}})Sun, Cheng, Gan, and Liu}]{DBLP:conf/emnlp/SunCGL19}
Siqi Sun, Yu~Cheng, Zhe Gan, and Jingjing Liu. 2019{\natexlab{a}}.
\newblock Patient knowledge distillation for {BERT} model compression.
\newblock In \emph{{EMNLP/IJCNLP} {(1)}}, pages 4322--4331. Association for Computational Linguistics.

\bibitem[{Sun et~al.(2019{\natexlab{b}})Sun, Deng, Nie, and Tang}]{DBLP:conf/iclr/SunDNT19}
Zhiqing Sun, Zhi{-}Hong Deng, Jian{-}Yun Nie, and Jian Tang. 2019{\natexlab{b}}.
\newblock Rotate: Knowledge graph embedding by relational rotation in complex space.
\newblock In \emph{{ICLR} (Poster)}. OpenReview.net.

\bibitem[{Tang et~al.(2019)Tang, Lu, Liu, Mou, Vechtomova, and Lin}]{DBLP:journals/corr/abs-1903-12136}
Raphael Tang, Yao Lu, Linqing Liu, Lili Mou, Olga Vechtomova, and Jimmy Lin. 2019.
\newblock Distilling task-specific knowledge from {BERT} into simple neural networks.
\newblock \emph{CoRR}, abs/1903.12136.

\bibitem[{Toutanova et~al.(2015)Toutanova, Chen, Pantel, Poon, Choudhury, and Gamon}]{DBLP:conf/emnlp/ToutanovaCPPCG15}
Kristina Toutanova, Danqi Chen, Patrick Pantel, Hoifung Poon, Pallavi Choudhury, and Michael Gamon. 2015.
\newblock Representing text for joint embedding of text and knowledge bases.
\newblock In \emph{{EMNLP}}, pages 1499--1509. The Association for Computational Linguistics.

\bibitem[{Trouillon et~al.(2016)Trouillon, Welbl, Riedel, Gaussier, and Bouchard}]{DBLP:conf/icml/TrouillonWRGB16}
Th{\'{e}}o Trouillon, Johannes Welbl, Sebastian Riedel, {\'{E}}ric Gaussier, and Guillaume Bouchard. 2016.
\newblock Complex embeddings for simple link prediction.
\newblock In \emph{{ICML}}, volume~48 of \emph{{JMLR} Workshop and Conference Proceedings}, pages 2071--2080. JMLR.org.

\bibitem[{Vashishth et~al.(2020)Vashishth, Sanyal, Nitin, and Talukdar}]{DBLP:conf/iclr/VashishthSNT20}
Shikhar Vashishth, Soumya Sanyal, Vikram Nitin, and Partha~P. Talukdar. 2020.
\newblock Composition-based multi-relational graph convolutional networks.
\newblock In \emph{{ICLR}}. OpenReview.net.

\bibitem[{Voita et~al.(2019)Voita, Talbot, Moiseev, Sennrich, and Titov}]{DBLP:conf/acl/VoitaTMST19}
Elena Voita, David Talbot, Fedor Moiseev, Rico Sennrich, and Ivan Titov. 2019.
\newblock Analyzing multi-head self-attention: Specialized heads do the heavy lifting, the rest can be pruned.
\newblock In \emph{{ACL} {(1)}}, pages 5797--5808. Association for Computational Linguistics.

\bibitem[{Wang et~al.(2019)Wang, Singh, Michael, Hill, Levy, and Bowman}]{DBLP:conf/iclr/WangSMHLB19}
Alex Wang, Amanpreet Singh, Julian Michael, Felix Hill, Omer Levy, and Samuel~R. Bowman. 2019.
\newblock \href {https://openreview.net/forum?id=rJ4km2R5t7} {{GLUE:} {A} multi-task benchmark and analysis platform for natural language understanding}.
\newblock In \emph{7th International Conference on Learning Representations, {ICLR} 2019, New Orleans, LA, USA, May 6-9, 2019}. OpenReview.net.

\bibitem[{Wang et~al.(2020{\natexlab{a}})Wang, Wu, Liu, Cai, Zhu, Gan, and Han}]{DBLP:conf/acl/WangWLCZGH20}
Hanrui Wang, Zhanghao Wu, Zhijian Liu, Han Cai, Ligeng Zhu, Chuang Gan, and Song Han. 2020{\natexlab{a}}.
\newblock {HAT:} hardware-aware transformers for efficient natural language processing.
\newblock In \emph{{ACL}}, pages 7675--7688. Association for Computational Linguistics.

\bibitem[{Wang et~al.(2021{\natexlab{a}})Wang, Wang, Lian, and Gao}]{DBLP:conf/cikm/WangWLG21}
Haoyu Wang, Yaqing Wang, Defu Lian, and Jing Gao. 2021{\natexlab{a}}.
\newblock A lightweight knowledge graph embedding framework for efficient inference and storage.
\newblock In \emph{{CIKM}}, pages 1909--1918. {ACM}.

\bibitem[{Wang et~al.(2021{\natexlab{b}})Wang, Liu, Ma, and Sheng}]{DBLP:conf/www/Wang0MS21}
Kai Wang, Yu~Liu, Qian Ma, and Quan~Z. Sheng. 2021{\natexlab{b}}.
\newblock Mulde: Multi-teacher knowledge distillation for low-dimensional knowledge graph embeddings.
\newblock In \emph{{WWW}}, pages 1716--1726. {ACM} / {IW3C2}.

\bibitem[{Wang et~al.(2020{\natexlab{b}})Wang, Wei, Dong, Bao, Yang, and Zhou}]{DBLP:conf/nips/WangW0B0020}
Wenhui Wang, Furu Wei, Li~Dong, Hangbo Bao, Nan Yang, and Ming Zhou. 2020{\natexlab{b}}.
\newblock \href {https://proceedings.neurips.cc/paper/2020/hash/3f5ee243547dee91fbd053c1c4a845aa-Abstract.html} {Minilm: Deep self-attention distillation for task-agnostic compression of pre-trained transformers}.
\newblock In \emph{Advances in Neural Information Processing Systems 33: Annual Conference on Neural Information Processing Systems 2020, NeurIPS 2020, December 6-12, 2020, virtual}.

\bibitem[{Wang et~al.(2014)Wang, Zhang, Feng, and Chen}]{DBLP:conf/aaai/WangZFC14}
Zhen Wang, Jianwen Zhang, Jianlin Feng, and Zheng Chen. 2014.
\newblock Knowledge graph embedding by translating on hyperplanes.
\newblock In \emph{{AAAI}}, pages 1112--1119. {AAAI} Press.

\bibitem[{Williams et~al.(2018)Williams, Nangia, and Bowman}]{DBLP:conf/naacl/WilliamsNB18}
Adina Williams, Nikita Nangia, and Samuel~R. Bowman. 2018.
\newblock \href {https://doi.org/10.18653/V1/N18-1101} {A broad-coverage challenge corpus for sentence understanding through inference}.
\newblock In \emph{Proceedings of the 2018 Conference of the North American Chapter of the Association for Computational Linguistics: Human Language Technologies, {NAACL-HLT} 2018, New Orleans, Louisiana, USA, June 1-6, 2018, Volume 1 (Long Papers)}, pages 1112--1122. Association for Computational Linguistics.

\bibitem[{Wolf et~al.(2020)Wolf, Debut, Sanh, Chaumond, Delangue, Moi, Cistac, Rault, Louf, Funtowicz, Davison, Shleifer, von Platen, Ma, Jernite, Plu, Xu, Scao, Gugger, Drame, Lhoest, and Rush}]{DBLP:conf/emnlp/WolfDSCDMCRLFDS20}
Thomas Wolf, Lysandre Debut, Victor Sanh, Julien Chaumond, Clement Delangue, Anthony Moi, Pierric Cistac, Tim Rault, R{\'{e}}mi Louf, Morgan Funtowicz, Joe Davison, Sam Shleifer, Patrick von Platen, Clara Ma, Yacine Jernite, Julien Plu, Canwen Xu, Teven~Le Scao, Sylvain Gugger, Mariama Drame, Quentin Lhoest, and Alexander~M. Rush. 2020.
\newblock \href {https://doi.org/10.18653/V1/2020.EMNLP-DEMOS.6} {Transformers: State-of-the-art natural language processing}.
\newblock In \emph{Proceedings of the 2020 Conference on Empirical Methods in Natural Language Processing: System Demonstrations, {EMNLP} 2020 - Demos, Online, November 16-20, 2020}, pages 38--45. Association for Computational Linguistics.

\bibitem[{Xu and Li(2019)}]{DBLP:conf/acl/XuL19}
Canran Xu and Ruijiang Li. 2019.
\newblock Relation embedding with dihedral group in knowledge graph.
\newblock In \emph{{ACL} {(1)}}, pages 263--272. Association for Computational Linguistics.

\bibitem[{Xu et~al.(2024)Xu, Wang, and Fan}]{DBLP:conf/coling/XuWF24}
Haotian Xu, Yuhua Wang, and Jiahui Fan. 2024.
\newblock Self-knowledge distillation for knowledge graph embedding.
\newblock In \emph{{LREC/COLING}}, pages 14595--14605. {ELRA} and {ICCL}.

\bibitem[{Yang et~al.(2015)Yang, Yih, He, Gao, and Deng}]{DBLP:journals/corr/YangYHGD14a}
Bishan Yang, Wen{-}tau Yih, Xiaodong He, Jianfeng Gao, and Li~Deng. 2015.
\newblock Embedding entities and relations for learning and inference in knowledge bases.
\newblock In \emph{{ICLR} (Poster)}.

\bibitem[{Zhang et~al.(2019)Zhang, Tay, Yao, and Liu}]{DBLP:conf/nips/0007TYL19}
Shuai Zhang, Yi~Tay, Lina Yao, and Qi~Liu. 2019.
\newblock Quaternion knowledge graph embeddings.
\newblock In \emph{NeurIPS}, pages 2731--2741.

\bibitem[{Zhang et~al.(2021)Zhang, Wong, Ye, Wen, Zhang, and Chen}]{DBLP:conf/icde/ZhangWYWZC21}
Wen Zhang, Chi~Man Wong, Ganqiang Ye, Bo~Wen, Wei Zhang, and Huajun Chen. 2021.
\newblock Billion-scale pre-trained e-commerce product knowledge graph model.
\newblock In \emph{{ICDE}}, pages 2476--2487. {IEEE}.

\bibitem[{Zhou et~al.(2022{\natexlab{a}})Zhou, Xu, and McAuley}]{DBLP:conf/acl/ZhouXM22}
Wangchunshu Zhou, Canwen Xu, and Julian~J. McAuley. 2022{\natexlab{a}}.
\newblock \href {https://doi.org/10.18653/V1/2022.ACL-LONG.485} {{BERT} learns to teach: Knowledge distillation with meta learning}.
\newblock In \emph{Proceedings of the 60th Annual Meeting of the Association for Computational Linguistics (Volume 1: Long Papers), {ACL} 2022, Dublin, Ireland, May 22-27, 2022}, pages 7037--7049. Association for Computational Linguistics.

\bibitem[{Zhou et~al.(2022{\natexlab{b}})Zhou, Chen, Wang, Feng, and Chen}]{DBLP:journals/corr/abs-2206-02963}
Zhehui Zhou, Defang Chen, Can Wang, Yan Feng, and Chun Chen. 2022{\natexlab{b}}.
\newblock Improving knowledge graph embedding via iterative self-semantic knowledge distillation.
\newblock \emph{CoRR}, abs/2206.02963.

\bibitem[{Zhu et~al.(2022)Zhu, Zhang, Chen, Chen, Cheng, Zhang, and Chen}]{DBLP:conf/wsdm/ZhuZCCC0C22}
Yushan Zhu, Wen Zhang, Mingyang Chen, Hui Chen, Xu~Cheng, Wei Zhang, and Huajun Chen. 2022.
\newblock Dualde: Dually distilling knowledge graph embedding for faster and cheaper reasoning.
\newblock In \emph{{WSDM}}, pages 1516--1524. {ACM}.

\bibitem[{Zhu et~al.(2021)Zhu, Zhao, Zhang, Ye, Chen, Zhang, and Chen}]{DBLP:conf/mm/ZhuZZYCZC21}
Yushan Zhu, Huaixiao Zhao, Wen Zhang, Ganqiang Ye, Hui Chen, Ningyu Zhang, and Huajun Chen. 2021.
\newblock Knowledge perceived multi-modal pretraining in e-commerce.
\newblock In \emph{{ACM} Multimedia}, pages 2744--2752. {ACM}.

\end{thebibliography}
\bibliographystyle{acl_natbib}

\appendix
\section{More Results of link prediction}
\label{sec:full_result}
\begin{figure*}[h]
	\centering  
  \subfigure[TransE on WN18RR]
  {\includegraphics[width=0.24\linewidth]{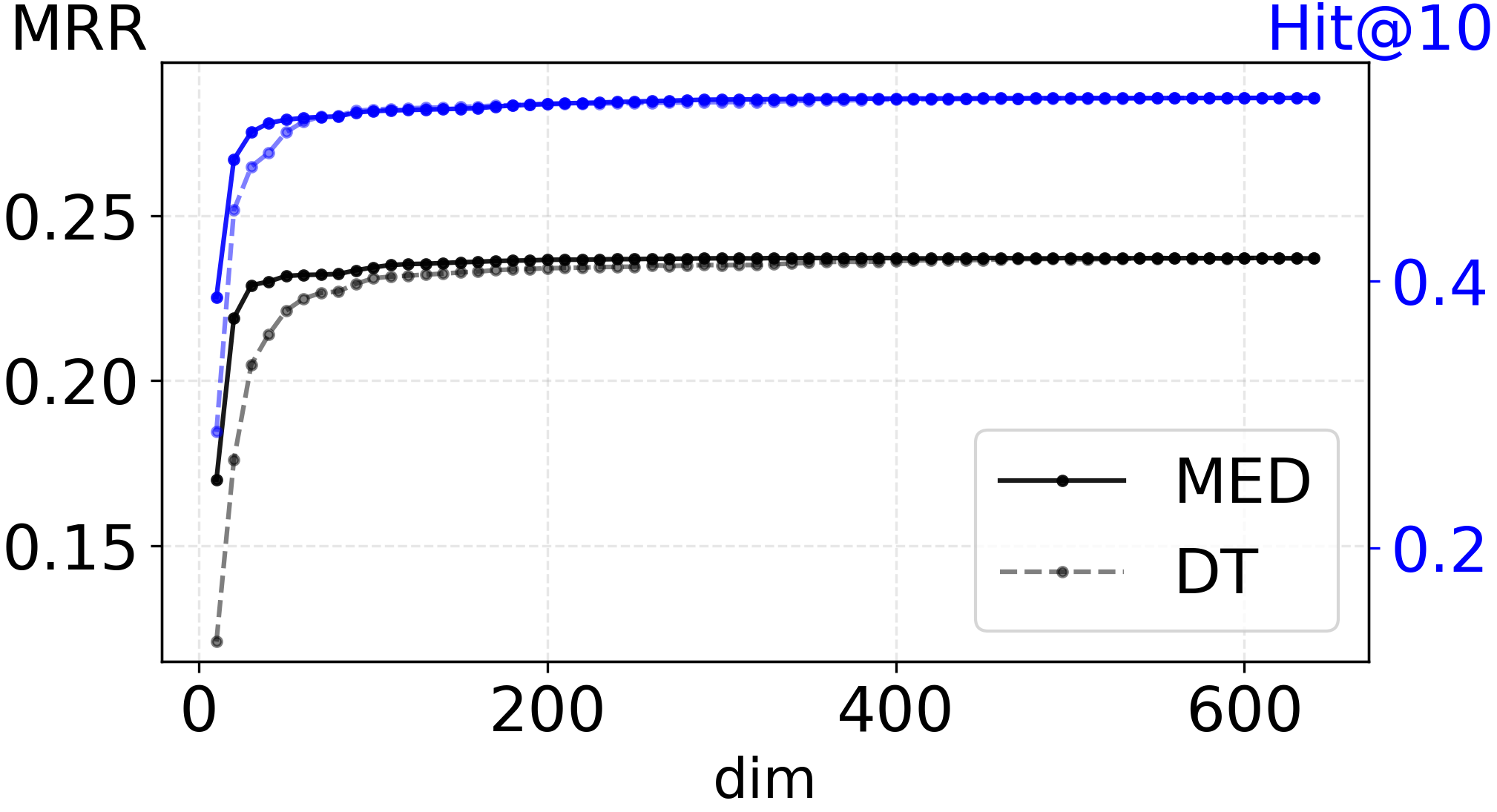}}
  \subfigure[SimplE on WN18RR]
  {\includegraphics[width=0.24\linewidth]{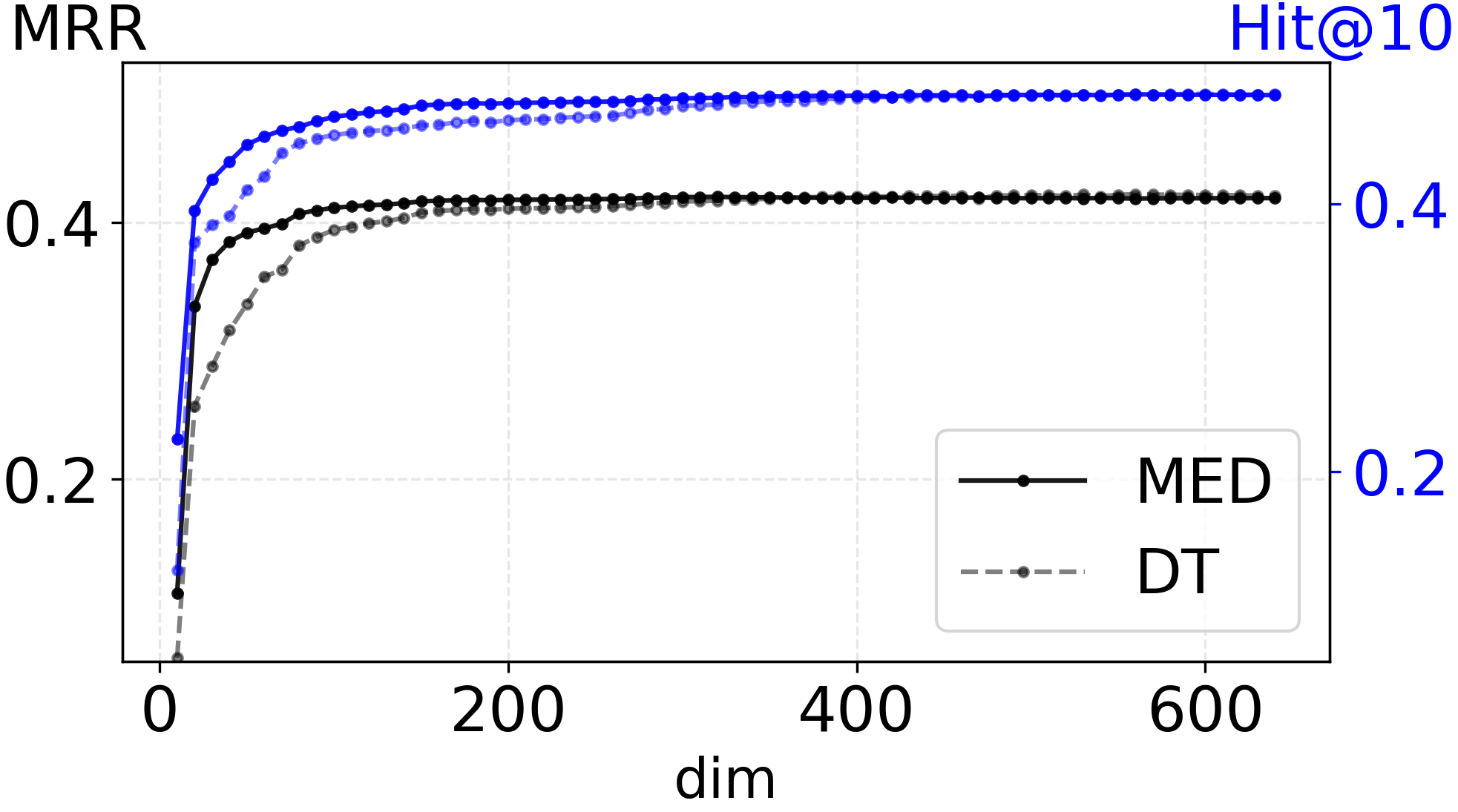}}
  \subfigure[RotatE on WN18RR]
  {\includegraphics[width=0.24\linewidth]{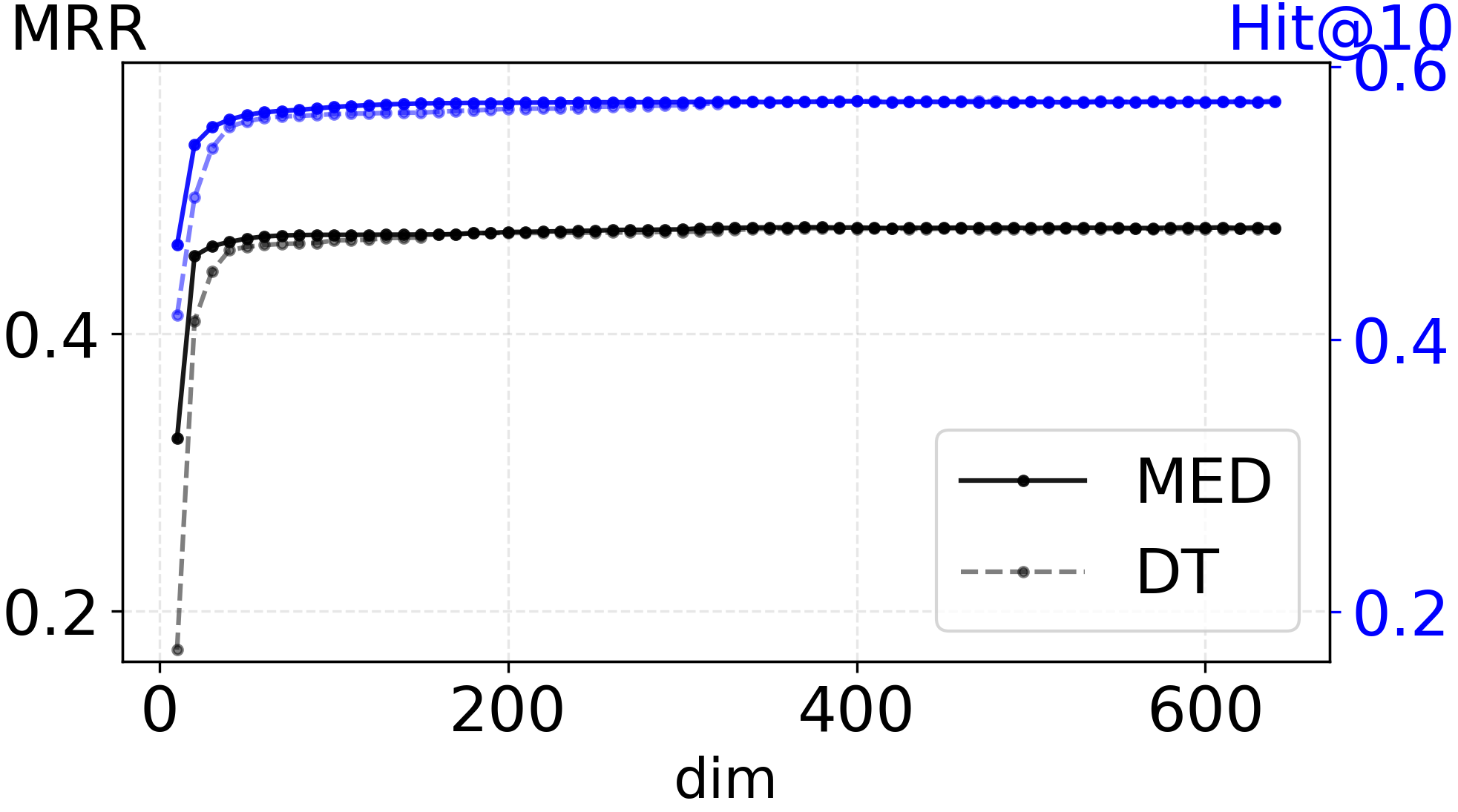}}
  \subfigure[PairRE on WN18RR]
  {\includegraphics[width=0.24\linewidth]{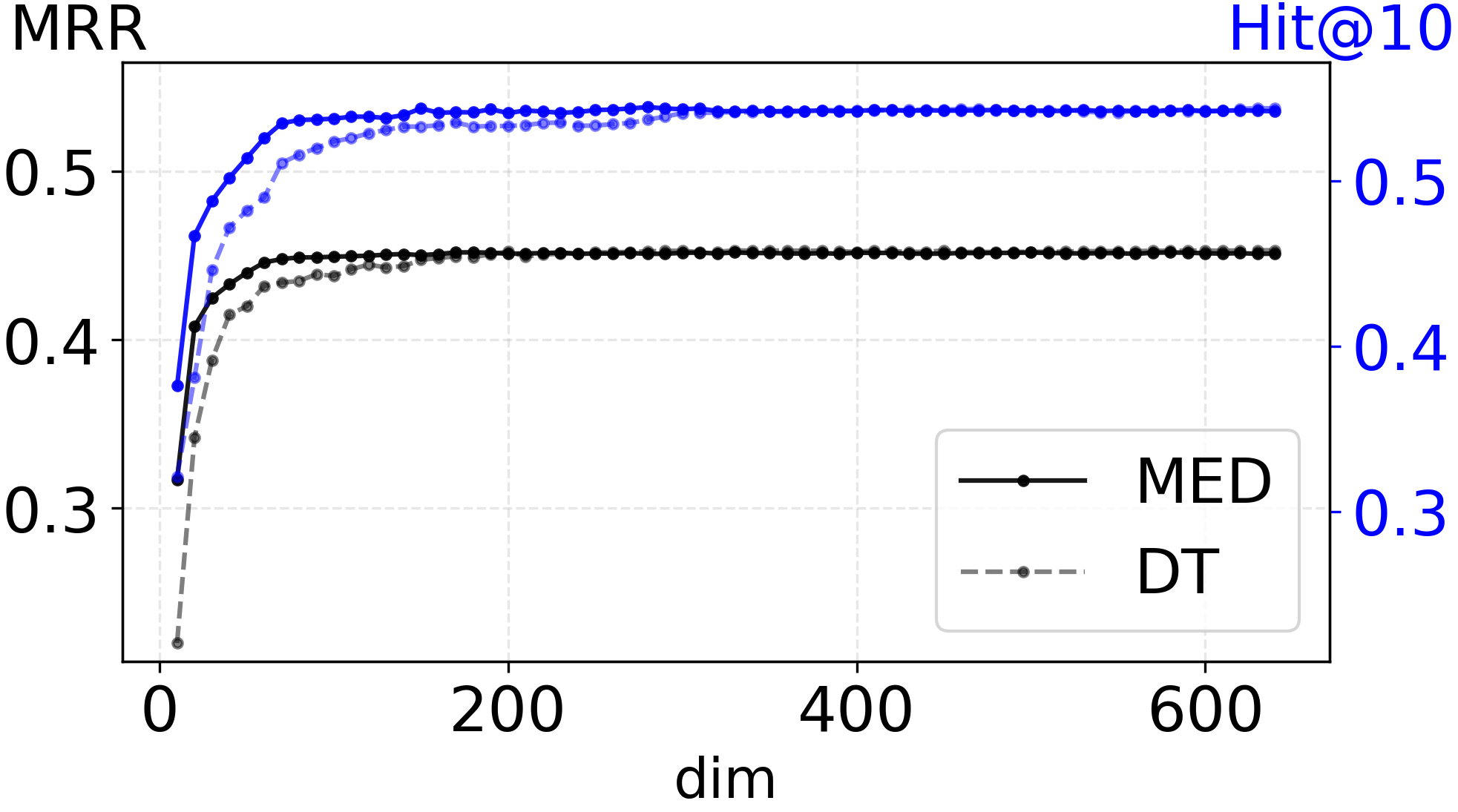}}

  \subfigure[TransE on FB15K237]
  {\includegraphics[width=0.24\linewidth]{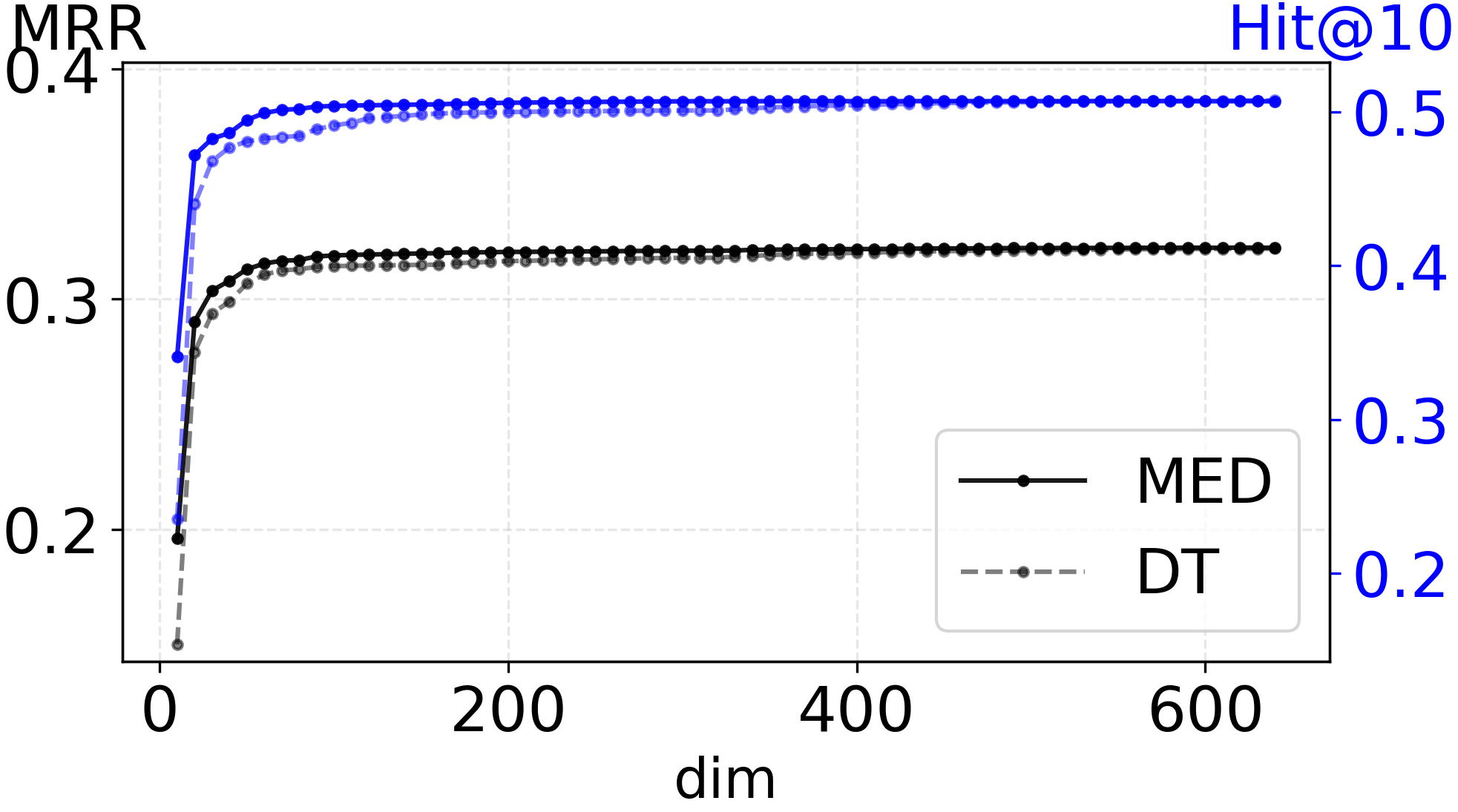}}
  \subfigure[SimplE on FB15K237]
  {\includegraphics[width=0.24\linewidth]{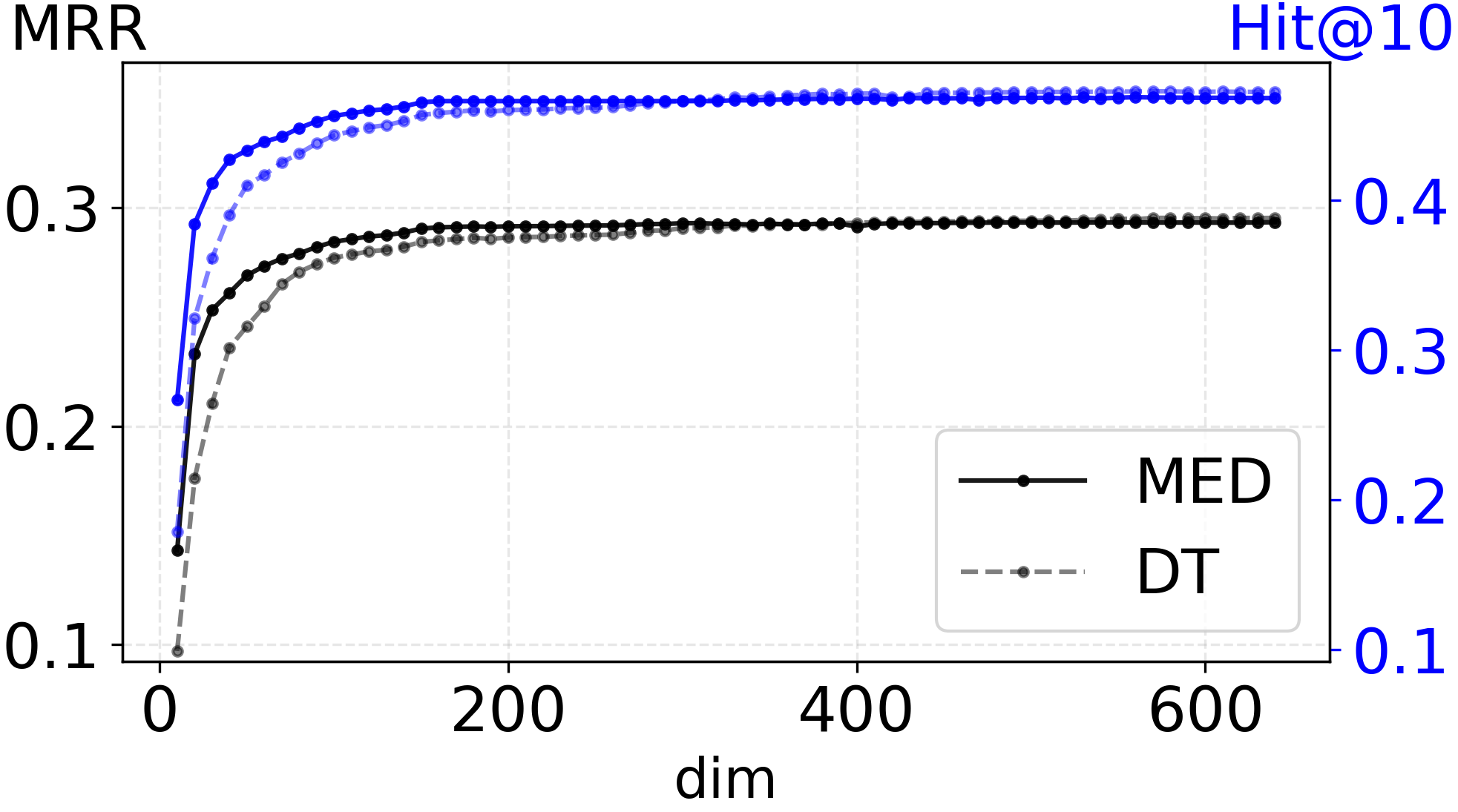}}
  \subfigure[RotatE on FB15K237]
  {\includegraphics[width=0.24\linewidth]{pic/FB15K237_rotate_MRR_Hit10.png}}
  \subfigure[PairRE on FB15K237]
  {\includegraphics[width=0.24\linewidth]{pic/FB15K237_pairre_MRR_Hit10.png}}
\caption{Performance of sub-models of {\model} and the directly trained (DT) KGEs of dimensions from 10 to 640.}
    \label{fig:all_dim}
\end{figure*}

More results of link prediction are shown in Table~\ref{table_wn_mrr_h1} and Table~\ref{table_wn_h10_h3} for WN18RR, and Table~\ref{table_fb_mrr_h1} and Table~\ref{table_fb_h10_h3} for FB15K237. All comparison results of sub-models of {\model} to the directly trained KGEs (DT) of 10- to 640-dimension are shown in Fig.~\ref{fig:all_dim}.

\setlength\tabcolsep{5pt}
\renewcommand\arraystretch{.9}
\begin{table*}[htp]

\begin{center}
        \resizebox{1.0\textwidth}{!}{
\begin{tabular}{cl|cc|cc|cc|cc|cc|cc|cc}
\toprule
\multicolumn{2}{l}{}                                & \multicolumn{2}{c}{\textbf{10d}} & \multicolumn{2}{c}{\textbf{20d}} & \multicolumn{2}{c}{\textbf{40d}} & \multicolumn{2}{c}{\textbf{80d}} & \multicolumn{2}{c}{\textbf{160d}} & \multicolumn{2}{c}{\textbf{320d}} & \multicolumn{2}{c}{\textbf{640d}} \\
\multirow{10}{*}{\textbf{TransE}} & \textit{Method} & \textit{MRR}   & \textit{Hit@10} & \textit{MRR}   & \textit{Hit@10} & \textit{MRR}   & \textit{Hit@10} & \textit{MRR}   & \textit{Hit@10} & \textit{MRR}    & \textit{Hit@10} & \textit{MRR}    & \textit{Hit@10} & \textit{MRR}    & \textit{Hit@10} \\
\midrule
                                  & DT              & 0.121          & 0.287           & 0.176          & 0.453           & 0.214          & 0.496           & 0.227          & 0.524           & 0.233           & 0.531           & 0.235           & 0.534           & \textbf{0.237}  & \textbf{0.537}  \\
                                  & Ext             & 0.125          & 0.298           & 0.172          & 0.423           & 0.199          & 0.468           & 0.213          & 0.495           & 0.225           & 0.515           & 0.226           & 0.521           & 0.237           & 0.537           \\
                                  & Ext-L           & 0.139          & 0.315           & 0.196          & 0.461           & 0.224          & 0.497           & 0.232          & 0.516           & 0.236           & \textbf{0.534}  & 0.236           & 0.535           & 0.237           & 0.537           \\
                                  & Ext-V           & 0.139          & 0.309           & 0.198          & 0.458           & 0.222          & 0.494           & 0.234          & 0.525           & 0.236           & 0.532           & 0.236           & 0.536           & 0.237           & 0.537           \\
                                  & BKD             & 0.141          & 0.323           & 0.207          & 0.480           & 0.226          & 0.513           & 0.232          & 0.527           & 0.233           & 0.531           & 0.236           & 0.533           & -               & -               \\
                                  & TA              & 0.144          & 0.335           & 0.211          & 0.483           & 0.226          & 0.512           & 0.233          & 0.527           & 0.234           & 0.533           & 0.236           & 0.535           & -               & -               \\
                                  & DualDE          & 0.148          & 0.337           & 0.213          & 0.488           & 0.225          & 0.514           & \textbf{0.234} & \textbf{0.530}  & 0.235           & 0.533           & \textbf{0.238}  & 0.535           & -               & -               \\
                                  & IterDE          & 0.143          & 0.332           & 0.211          & 0.484           & 0.224          & 0.511           & 0.232          & 0.528           & 0.236           & 0.531           & 0.237           & 0.533           & -               & -               \\
                                  & {\model}        & \textbf{0.170} & \textbf{0.388}  & \textbf{0.219} & \textbf{0.491}  & \textbf{0.232} & \textbf{0.518}  & 0.232          & 0.523           & \textbf{0.236}  & 0.529           & 0.237           & \textbf{0.536}  & 0.237           & 0.537           \\
                                  \midrule
\multirow{10}{*}{\textbf{SimplE}} & \textit{Method} & \textit{MRR}   & \textit{Hit@10} & \textit{MRR}   & \textit{Hit@10} & \textit{MRR}   & \textit{Hit@10} & \textit{MRR}   & \textit{Hit@10} & \textit{MRR}    & \textit{Hit@10} & \textit{MRR}    & \textit{Hit@10} & \textit{MRR}    & \textit{Hit@10} \\
                                  & DT              & 0.061          & 0.126           & 0.257          & 0.372           & 0.316          & 0.389           & 0.382          & 0.446           & 0.409           & 0.459           & 0.417           & 0.474           & \textbf{0.421}  & 0.481           \\
                                  & Ext             & 0.004          & 0.007           & 0.051          & 0.107           & 0.160          & 0.249           & 0.219          & 0.314           & 0.357           & 0.401           & 0.407           & 0.451           & 0.421           & 0.481           \\
                                  & Ext-L           & 0.005          & 0.006           & 0.048          & 0.078           & 0.169          & 0.244           & 0.369          & 0.435           & 0.398           & 0.454           & 0.417           & 0.481           & 0.421           & 0.481           \\
                                  & Ext-V           & 0.004          & 0.006           & 0.047          & 0.076           & 0.246          & 0.317           & 0.368          & 0.402           & 0.398           & 0.461           & 0.413           & 0.472           & 0.421           & 0.481           \\
                                  & BKD             & 0.075          & 0.156           & 0.285          & 0.381           & 0.343          & 0.399           & 0.394          & 0.450           & 0.414           & 0.468           & 0.418           & 0.475           & -               & -               \\
                                  & TA              & 0.089          & 0.189           & 0.316          & 0.386           & 0.368          & 0.418           & 0.405          & 0.456           & 0.415           & 0.472           & 0.421           & 0.481           & -               & -               \\
                                  & DualDE          & 0.083          & 0.175           & 0.328          & 0.388           & \textbf{0.386} & 0.423           & 0.407          & 0.454           & \textbf{0.419}  & 0.475           & \textbf{0.422}  & \textbf{0.482}  & -               & -               \\
                                  & IterDE          & 0.077          & 0.162           & 0.321          & 0.378           & 0.375          & 0.419           & 0.404          & 0.452           & 0.416           & 0.469           & 0.421           & 0.482           & -               & -               \\
                                  & {\model}        & \textbf{0.111} & \textbf{0.224}  & \textbf{0.335} & \textbf{0.395}  & 0.385          & \textbf{0.431}  & \textbf{0.407} & \textbf{0.457}  & 0.418           & \textbf{0.477}  & 0.421           & 0.481           & 0.421           & \textbf{0.482}  \\
                                  \midrule
\multirow{10}{*}{\textbf{RotatE}} & \textit{Method} & \textit{MRR}   & \textit{Hit@10} & \textit{MRR}   & \textit{Hit@10} & \textit{MRR}   & \textit{Hit@10} & \textit{MRR}   & \textit{Hit@10} & \textit{MRR}    & \textit{Hit@10} & \textit{MRR}    & \textit{Hit@10} & \textit{MRR}    & \textit{Hit@10} \\
                                  & DT              & 0.172          & 0.418           & 0.409          & 0.504           & 0.456          & 0.556           & 0.465          & 0.564           & 0.471           & 0.567           & 0.474           & 0.573           & \textbf{0.476}  & \textbf{0.575}  \\
                                  & Ext             & 0.299          & 0.378           & 0.379          & 0.464           & 0.437          & 0.516           & 0.458          & 0.544           & 0.467           & 0.549           & 0.471           & 0.552           & 0.476           & 0.575           \\
                                  & Ext-L           & 0.206          & 0.277           & 0.336          & 0.424           & 0.399          & 0.487           & 0.423          & 0.515           & 0.445           & \textbf{0.541}  & 0.466           & 0.564           & 0.476           & 0.575           \\
                                  & Ext-V           & 0.261          & 0.377           & 0.304          & 0.433           & 0.337          & 0.471           & 0.366          & 0.497           & 0.416           & 0.532           & 0.451           & 0.561           & 0.476           & 0.575           \\
                                  & BKD             & 0.175          & 0.434           & 0.424          & 0.540           & 0.457          & 0.556           & 0.471          & 0.565           & 0.472           & 0.570           & 0.474           & 0.572           & -               & -               \\
                                  & TA              & 0.177          & 0.438           & 0.424          & 0.542           & 0.459          & 0.558           & 0.470          & 0.567           & 0.473           & 0.572           & 0.474           & 0.572           & -               & -               \\
                                  & DualDE          & 0.179          & 0.440           & 0.425          & 0.542           & 0.462          & 0.559           & \textbf{0.471} & 0.567           & \textbf{0.473}  & 0.573           & 0.475           & 0.573           & -               & -               \\
                                  & IterDE          & 0.176          & 0.436           & 0.421          & 0.538           & 0.459          & 0.560           & 0.470          & 0.567           & 0.471           & 0.569           & 0.474           & 0.572           & -               & -               \\
                                  & {\model}        & \textbf{0.324} & \textbf{0.469}  & \textbf{0.456} & \textbf{0.543}  & \textbf{0.466} & \textbf{0.561}  & 0.471          & \textbf{0.568}  & 0.471           & \textbf{0.574}  & \textbf{0.476}  & \textbf{0.573}  & 0.476           & 0.574           \\
                                  \midrule
\multirow{10}{*}{\textbf{PairRE}} & \textit{Method} & \textit{MRR}   & \textit{Hit@10} & \textit{MRR}   & \textit{Hit@10} & \textit{MRR}   & \textit{Hit@10} & \textit{MRR}   & \textit{Hit@10} & \textit{MRR}    & \textit{Hit@10} & \textit{MRR}    & \textit{Hit@10} & \textit{MRR}    & \textit{Hit@10} \\
                                  & DT              & 0.220          & 0.321           & 0.342          & 0.381           & 0.415          & 0.472           & 0.435          & 0.516           & 0.449           & 0.534           & 0.452           & 0.542           & \textbf{0.453}  & \textbf{0.544}  \\
                                  & Ext             & 0.152          & 0.209           & 0.261          & 0.379           & 0.334          & 0.463           & 0.375          & 0.493           & 0.419           & 0.526           & 0.438           & 0.545           & 0.453           & 0.544           \\
                                  & Ext-L           & 0.162          & 0.220           & 0.281          & 0.360           & 0.363          & 0.442           & 0.417          & 0.495           & 0.437           & 0.523           & 0.446           & 0.544           & 0.453           & 0.544           \\
                                  & Ext-V           & 0.172          & 0.260           & 0.306          & 0.374           & 0.389          & 0.456           & 0.420          & 0.498           & 0.441           & 0.529           & 0.446           & 0.541           & 0.453           & 0.544           \\
                                  & BKD             & 0.228          & 0.336           & 0.375          & 0.413           & 0.421          & 0.483           & 0.443          & 0.525           & 0.451           & 0.536           & 0.453           & 0.542           & -               & -               \\
                                  & TA              & 0.245          & 0.340           & 0.381          & 0.427           & 0.426          & 0.487           & 0.448          & 0.534           & 0.452           & 0.537           & 0.453           & 0.543           & -               & -               \\
                                  & DualDE          & 0.242          & 0.336           & 0.377          & 0.424           & 0.428          & 0.495           & \textbf{0.451} & 0.536           & \textbf{0.453}  & 0.540           & \textbf{0.454}  & \textbf{0.544}  & -               & -               \\
                                  & IterDE          & 0.235          & 0.336           & 0.379          & 0.423           & 0.426          & 0.495           & 0.449          & 0.533           & 0.450           & 0.538           & 0.452           & 0.543           & -               & -               \\
                                  & {\model}        & \textbf{0.317} & \textbf{0.376}  & \textbf{0.408} & \textbf{0.467}  & \textbf{0.433} & \textbf{0.502}  & 0.449          & \textbf{0.537}  & 0.451           & \textbf{0.541}  & 0.451           & 0.542           & 0.451           & 0.542          \\
                                  \bottomrule
\end{tabular}
}
\end{center}
\caption{MRR and Hit@10 of some representative dimensions on WN18RR.}
\label{table_wn_mrr_h1}
\end{table*}

\setlength\tabcolsep{4pt}
\renewcommand\arraystretch{.9}
\begin{table*}[htp]

\begin{center}
    \resizebox{1.0\textwidth}{!}{
\begin{tabular}{cl|cc|cc|cc|cc|cc|cc|cc}
\toprule
\multicolumn{2}{l}{}                                & \multicolumn{2}{c}{\textbf{10d}} & \multicolumn{2}{c}{\textbf{20d}} & \multicolumn{2}{c}{\textbf{40d}} & \multicolumn{2}{c}{\textbf{80d}} & \multicolumn{2}{c}{\textbf{160d}} & \multicolumn{2}{c}{\textbf{320d}} & \multicolumn{2}{c}{\textbf{640d}} \\
\midrule
\multirow{10}{*}{\textbf{TransE}} & \textit{Method} & \textit{Hit@3}  & \textit{Hit@1} & \textit{Hit@3}  & \textit{Hit@1} & \textit{Hit@3}  & \textit{Hit@1} & \textit{Hit@3}  & \textit{Hit@1} & \textit{Hit@3}  & \textit{Hit@1}  & \textit{Hit@3}  & \textit{Hit@1}  & \textit{Hit@3}  & \textit{Hit@1}  \\
                                  & DT              & 0.202           & 0.011          & 0.291           & 0.016          & 0.385           & 0.018          & 0.401           & 0.025          & 0.403           & 0.027           & 0.407           & 0.033           & 0.412           & \textbf{0.034}  \\
                                  & Ext             & 0.201           & 0.016          & 0.285           & 0.023          & 0.338           & 0.023          & 0.364           & 0.028          & 0.384           & 0.033           & 0.388           & 0.028           & 0.412           & 0.034           \\
                                  & Ext-L           & 0.218           & 0.029          & 0.317           & 0.025          & 0.361           & 0.039          & 0.403           & 0.046          & 0.405           & 0.036           & 0.408           & 0.033           & 0.412           & 0.034           \\
                                  & Ext-V           & 0.218           & 0.029          & 0.314           & \textbf{0.045} & 0.391           & \textbf{0.051} & 0.407           & \textbf{0.047} & 0.408           & 0.036           & 0.411           & 0.027           & 0.412           & 0.034           \\
                                  & BKD             & 0.216           & 0.035          & 0.331           & 0.040          & 0.392           & 0.033          & 0.401           & 0.031          & 0.404           & 0.030           & 0.407           & 0.032           & -               & -               \\
                                  & TA              & 0.224           & 0.040          & 0.343           & 0.043          & 0.395           & 0.037          & 0.408           & 0.030          & 0.407           & 0.030           & 0.410           & 0.034           & -               & -               \\
                                  & DualDE          & 0.226           & 0.037          & 0.346           & 0.043          & 0.394           & 0.037          & \textbf{0.408}  & 0.031          & \textbf{0.408}  & 0.031           & \textbf{0.411}  & \textbf{0.034}  & -               & -               \\
                                  & IterDE          & 0.217           & 0.032          & 0.345           & 0.044          & 0.392           & 0.036          & 0.407           & 0.030          & 0.408           & 0.031           & 0.407           & 0.033           & -               & -               \\
                                  & {\model}        & \textbf{0.269}  & \textbf{0.040} & \textbf{0.369}  & \textbf{0.045} & \textbf{0.399}  & 0.038          & 0.404           & 0.042          & 0.407           & \textbf{0.037}  & 0.410           & 0.033           & \textbf{0.412}  & 0.031           \\
                                  \midrule
\multirow{10}{*}{\textbf{SimplE}} & \textit{Method} & \textit{Hit@3}  & \textit{Hit@1} & \textit{Hit@3}  & \textit{Hit@1} & \textit{Hit@3}  & \textit{Hit@1} & \textit{Hit@3}  & \textit{Hit@1} & \textit{Hit@3}  & \textit{Hit@1}  & \textit{Hit@3}  & \textit{Hit@1}  & \textit{Hit@3}  & \textit{Hit@1}  \\
                                  & DT              & 0.061           & 0.028          & 0.297           & 0.193          & 0.361           & 0.289          & 0.406           & 0.343          & 0.420           & 0.382           & 0.428           & 0.386           & 0.433           & \textbf{0.391}  \\
                                  & Ext             & 0.003           & 0.001          & 0.055           & 0.023          & 0.181           & 0.114          & 0.249           & 0.168          & 0.377           & 0.329           & 0.422           & 0.381           & 0.433           & 0.391           \\
                                  & Ext-L           & 0.004           & 0.003          & 0.051           & 0.031          & 0.187           & 0.128          & 0.389           & 0.333          & 0.413           & 0.365           & 0.429           & 0.384           & 0.433           & 0.391           \\
                                  & Ext-V           & 0.004           & 0.002          & 0.050           & 0.029          & 0.269           & 0.205          & 0.378           & 0.349          & 0.409           & 0.372           & 0.426           & 0.382           & 0.433           & 0.391           \\
                                  & BKD             & 0.077           & 0.034          & 0.331           & 0.225          & 0.384           & 0.311          & 0.415           & 0.358          & 0.426           & 0.371           & 0.431           & 0.385           & -               & -               \\
                                  & TA              & 0.093           & 0.042          & 0.349           & 0.269          & 0.375           & 0.349          & 0.412           & \textbf{0.384} & 0.425           & 0.388           & 0.431           & 0.389           & -               & -               \\
                                  & DualDE          & 0.086           & 0.038          & 0.361           & 0.285          & 0.391           & \textbf{0.368} & 0.416           & 0.383          & 0.427           & \textbf{0.389}  & 0.434           & \textbf{0.392}  & -               & -               \\
                                  & IterDE          & 0.079           & 0.033          & 0.355           & 0.279          & 0.382           & 0.356          & 0.415           & 0.379          & 0.424           & 0.383           & 0.433           & 0.389           & \textbf{-}      & \textbf{-}      \\
                                  & {\model}        & \textbf{0.119}  & \textbf{0.048} & \textbf{0.366}  & \textbf{0.292} & \textbf{0.395}  & 0.359          & \textbf{0.419}  & 0.380          & \textbf{0.429}  & 0.389           & \textbf{0.435}  & 0.391           & \textbf{0.434}  & 0.390           \\
                                  \midrule
\multirow{10}{*}{\textbf{RotatE}} & \textit{Method} & \textit{Hit@3}  & \textit{Hit@1} & \textit{Hit@3}  & \textit{Hit@1} & \textit{Hit@3}  & \textit{Hit@1} & \textit{Hit@3}  & \textit{Hit@1} & \textit{Hit@3}  & \textit{Hit@1}  & \textit{Hit@3}  & \textit{Hit@1}  & \textit{Hit@3}  & \textit{Hit@1}  \\
                                  & DT              & 0.304           & 0.005          & 0.436           & 0.357          & 0.475           & 0.393          & 0.487           & 0.420          & 0.489           & 0.423           & 0.491           & 0.428           & 0.493           & \textbf{0.429}  \\
                                  & Ext             & 0.315           & 0.257          & 0.399           & 0.335          & 0.452           & 0.395          & 0.472           & 0.415          & 0.480           & 0.413           & 0.470           & 0.418           & 0.493           & 0.429           \\
                                  & Ext-L           & 0.224           & 0.166          & 0.359           & 0.288          & 0.420           & 0.352          & 0.441           & 0.373          & 0.461           & 0.396           & 0.481           & 0.417           & 0.493           & 0.429           \\
                                  & Ext-V           & 0.289           & 0.197          & 0.336           & 0.234          & 0.377           & 0.263          & 0.402           & 0.293          & 0.442           & 0.357           & 0.467           & 0.397           & 0.493           & 0.429           \\
                                  & BKD             & 0.312           & 0.009          & 0.452           & 0.361          & 0.479           & 0.403          & 0.487           & 0.421          & 0.490           & 0.424           & 0.492           & 0.425           & -               & -               \\
                                  & TA              & 0.314           & 0.010          & 0.452           & 0.363          & 0.481           & 0.408          & 0.489           & 0.420          & 0.488           & 0.422           & 0.492           & 0.425           & -               & -               \\
                                  & DualDE          & 0.320           & 0.011          & 0.452           & 0.364          & 0.483           & 0.412          & 0.489           & \textbf{0.423} & 0.488           & \textbf{0.426}  & 0.491           & 0.425           & -               & -               \\
                                  & IterDE          & 0.311           & 0.013          & 0.439           & 0.356          & 0.479           & 0.407          & 0.484           & 0.423          & 0.488           & 0.425           & 0.493           & 0.424           & -               & -               \\
                                  & {\model}        & \textbf{0.354}  & \textbf{0.277} & \textbf{0.476}  & \textbf{0.409} & \textbf{0.486}  & \textbf{0.418} & \textbf{0.490}  & 0.422          & \textbf{0.492}  & 0.424           & \textbf{0.493}  & \textbf{0.427}  & \textbf{0.495}  & 0.428           \\
                                  \midrule
\multirow{10}{*}{\textbf{PairRE}} & \textit{Method} & \textit{Hit@3}  & \textit{Hit@1} & \textit{Hit@3}  & \textit{Hit@1} & \textit{Hit@3}  & \textit{Hit@1} & \textit{Hit@3}  & \textit{Hit@1} & \textit{Hit@3}  & \textit{Hit@1}  & \textit{Hit@3}  & \textit{Hit@1}  & \textit{Hit@3}  & \textit{Hit@1}  \\
                                  & DT              & 0.271           & 0.174         & 0.368           & 0.313          & 0.428           & 0.384          & 0.450           & 0.399          & 0.463           & 0.405           & 0.462           & 0.406           & \textbf{0.464}  & \textbf{0.407}  \\
                                  & Ext             & 0.163           & 0.120          & 0.292           & 0.198          & 0.366           & 0.267          & 0.398           & 0.314          & 0.437           & 0.364           & 0.452           & 0.388           & 0.464           & 0.407           \\
                                  & Ext-L           & 0.175           & 0.129          & 0.302           & 0.237          & 0.383           & 0.319          & 0.431           & 0.377          & 0.450           & 0.395           & 0.455           & 0.400           & 0.464           & 0.407           \\
                                  & Ext-V           & 0.192           & 0.124          & 0.323           & 0.269          & 0.407           & 0.352          & 0.435           & 0.379          & 0.452           & 0.398           & 0.458           & 0.400           & 0.464           & 0.407           \\
                                  & BKD             & 0.279           & 0.184          & 0.388           & 0.334          & 0.435           & 0.372          & 0.452           & 0.405          & 0.460           & 0.405           & 0.463           & 0.407           & -               & -               \\
                                  & TA              & 0.293           & 0.197          & 0.387           & 0.332          & 0.437           & 0.380          & 0.460           & 0.404          & 0.462           & 0.409           & 0.463           & 0.408           & -               & -               \\
                                  & DualDE          & 0.281           & 0.175          & 0.389           & 0.330          & 0.437           & 0.381          & \textbf{0.463}  & \textbf{0.409} & 0.463           & 0.410           & 0.465           & \textbf{0.410}  & -               & -               \\
                                  & IterDE          & 0.285           & 0.172          & 0.390           & 0.331          & 0.435           & 0.377          & 0.461           & 0.405          & 0.463           & \textbf{0.411}  & 0.464           & 0.410           & -               & -               \\
                                  & {\model}        & \textbf{0.314}  & \textbf{0.259} & \textbf{0.426}  & \textbf{0.367} & \textbf{0.443}  & \textbf{0.392} & 0.462           & 0.405          & \textbf{0.464}  & 0.406           & \textbf{0.465}  & 0.407           & 0.464           & 0.406          \\
                                  \bottomrule
\end{tabular}
}
\end{center}
\caption{Hit@3 and Hit@1 of some representative dimensions on WN18RR.}
\label{table_wn_h10_h3}
\end{table*}

\setlength\tabcolsep{5pt}
\renewcommand\arraystretch{.9}
\begin{table*}[htp]
\begin{center}
    \resizebox{1.0\textwidth}{!}{
\begin{tabular}{cl|cc|cc|cc|cc|cc|cc|cc}
\toprule
\multicolumn{2}{l}{}                                & \multicolumn{2}{c}{\textbf{10d}} & \multicolumn{2}{c}{\textbf{20d}} & \multicolumn{2}{c}{\textbf{40d}} & \multicolumn{2}{c}{\textbf{80d}} & \multicolumn{2}{c}{\textbf{160d}} & \multicolumn{2}{c}{\textbf{320d}} & \multicolumn{2}{c}{\textbf{640d}} \\
\midrule
\multirow{10}{*}{\textbf{TransE}} & \textit{Method} & \textit{MRR}   & \textit{Hit@10} & \textit{MRR}   & \textit{Hit@10} & \textit{MRR}   & \textit{Hit@10} & \textit{MRR}   & \textit{Hit@10} & \textit{MRR}    & \textit{Hit@10} & \textit{MRR}    & \textit{Hit@10} & \textit{MRR}    & \textit{Hit@10} \\
                                  & DT              & 0.150          & 0.235           & 0.277          & 0.440           & 0.299          & 0.477           & 0.313          & 0.484           & 0.315           & 0.499           & 0.318           & 0.501           & 0.322           & \textbf{0.508}  \\
                                  & Ext             & 0.115          & 0.211           & 0.191          & 0.324           & 0.236          & 0.392           & 0.266          & 0.436           & 0.286           & 0.462           & 0.299           & 0.479           & 0.322           & 0.508           \\
                                  & Ext-L           & 0.109          & 0.194           & 0.175          & 0.293           & 0.232          & 0.381           & 0.263          & 0.424           & 0.285           & \textbf{0.462}  & 0.301           & 0.484           & 0.322           & 0.508           \\
                                  & Ext-V           & 0.139          & 0.256           & 0.200          & 0.348           & 0.237          & 0.396           & 0.270          & 0.437           & 0.293           & 0.466           & 0.308           & 0.488           & 0.322           & 0.508           \\
                                  & BKD             & 0.176          & 0.293           & 0.279          & 0.446           & 0.303          & 0.480           & 0.315          & 0.500           & 0.315           & 0.501           & 0.320           & 0.502           & -               & -               \\
                                  & TA              & 0.175          & 0.246           & 0.281          & 0.441           & 0.303          & 0.484           & 0.314          & 0.498           & 0.319           & 0.504           & 0.321           & 0.504           & -               & -               \\
                                  & DualDE          & 0.179          & 0.301           & 0.281          & 0.443           & 0.306          & 0.483           & 0.316          & 0.502           & 0.319           & 0.505           & \textbf{0.322}  & \textbf{0.508}  & -               & -               \\
                                  & IterDE          & 0.176          & 0.285           & 0.276          & 0.446           & 0.307          & 0.482           & 0.315          & \textbf{0.503}  & 0.317           & 0.505           & 0.319           & 0.505           & -               & -               \\
                                  & {\model}        & \textbf{0.196} & \textbf{0.341}  & \textbf{0.290} & \textbf{0.472}  & \textbf{0.308} & \textbf{0.486}  & \textbf{0.317} & 0.502           & \textbf{0.320}  & \textbf{0.505}  & 0.321           & 0.507           & \textbf{0.322}  & 0.507           \\
                                  \midrule
\multirow{10}{*}{\textbf{SimplE}} & \textit{Method} & \textit{MRR}   & \textit{Hit@10} & \textit{MRR}   & \textit{Hit@10} & \textit{MRR}   & \textit{Hit@10} & \textit{MRR}   & \textit{Hit@10} & \textit{MRR}    & \textit{Hit@10} & \textit{MRR}    & \textit{Hit@10} & \textit{MRR}    & \textit{Hit@10} \\
                                  & DT              & 0.097          & 0.179           & 0.176          & 0.321           & 0.236          & 0.390           & 0.271          & 0.431           & 0.285           & 0.458           & 0.291           & 0.467           & \textbf{0.295}  & \textbf{0.472}  \\
                                  & Ext             & 0.037          & 0.068           & 0.069          & 0.107           & 0.090          & 0.144           & 0.159          & 0.258           & 0.229           & 0.372           & 0.269           & 0.432           & 0.295           & 0.472           \\
                                  & Ext-L           & 0.045          & 0.059           & 0.056          & 0.062           & 0.083          & 0.146           & 0.114          & 0.205           & 0.196           & 0.316           & 0.258           & 0.421           & 0.295           & 0.472           \\
                                  & Ext-V           & 0.049          & 0.069           & 0.066          & 0.101           & 0.105          & 0.149           & 0.138          & 0.224           & 0.224           & 0.369           & 0.261           & 0.414           & 0.295           & 0.472           \\
                                  & BKD             & 0.113          & 0.204           & 0.182          & 0.315           & 0.244          & 0.412           & 0.275          & 0.439           & 0.287           & 0.463           & 0.293           & 0.470           & -               & -               \\
                                  & TA              & 0.124          & 0.221           & 0.192          & 0.329           & 0.254          & 0.416           & 0.276          & 0.448           & 0.290           & 0.465           & 0.295           & 0.471           & -               & -               \\
                                  & DualDE          & 0.120          & 0.213           & 0.195          & 0.346           & 0.258          & \textbf{0.429}  & 0.279          & 0.443           & \textbf{0.293}  & 0.466           & \textbf{0.296}  & \textbf{0.468}  & -               & -               \\
                                  & IterDE          & 0.120          & 0.215           & 0.193          & 0.338           & 0.257          & 0.427           & 0.281          & 0.440           & 0.293           & 0.465           & 0.297           & 0.468           & -               & -               \\
                                  & {\model}        & \textbf{0.143} & \textbf{0.267}  & \textbf{0.233} & \textbf{0.384}  & \textbf{0.261} & 0.427           & \textbf{0.279} & \textbf{0.448}  & 0.291           & \textbf{0.466}  & 0.293           & 0.468           & 0.294           & 0.470           \\
                                  \midrule
\multirow{10}{*}{\textbf{RotatE}} & \textit{Method} & \textit{MRR}   & \textit{Hit@10} & \textit{MRR}   & \textit{Hit@10} & \textit{MRR}   & \textit{Hit@10} & \textit{MRR}   & \textit{Hit@10} & \textit{MRR}    & \textit{Hit@10} & \textit{MRR}    & \textit{Hit@10} & \textit{MRR}    & \textit{Hit@10} \\
                                  & DT              & 0.254          & 0.424           & 0.297          & 0.477           & 0.312          & 0.495           & 0.317          & 0.502           & 0.322           & 0.506           & 0.323           & 0.510           & \textbf{0.325}  & \textbf{0.515}  \\
                                  & Ext             & 0.138          & 0.245           & 0.203          & 0.340           & 0.251          & 0.410           & 0.276          & 0.443           & 0.291           & 0.465           & 0.305           & 0.485           & 0.325           & 0.515           \\
                                  & Ext-L           & 0.135          & 0.243           & 0.188          & 0.319           & 0.221          & 0.365           & 0.246          & 0.402           & 0.280           & 0.453           & 0.299           & 0.477           & 0.325           & 0.515           \\
                                  & Ext-V           & 0.160          & 0.281           & 0.198          & 0.340           & 0.238          & 0.393           & 0.265          & 0.427           & 0.288           & 0.458           & 0.302           & 0.478           & 0.325           & 0.515           \\
                                  & BKD             & 0.277          & 0.442           & 0.305          & 0.485           & 0.314          & 0.503           & 0.321          & 0.508           & 0.322           & 0.510           & 0.323           & 0.509           & -               & -               \\
                                  & TA              & 0.280          & 0.447           & 0.306          & 0.485           & 0.313          & 0.501           & 0.319          & 0.507           & 0.323           & 0.510           & 0.323           & 0.509           & -               & -               \\
                                  & DualDE          & 0.282          & 0.449           & 0.307          & 0.486           & 0.315          & 0.502           & 0.318          & 0.507           & 0.322           & \textbf{0.512}  & 0.324           & \textbf{0.514}  & -               & -               \\
                                  & IterDE          & 0.276          & 0.445           & 0.306          & 0.482           & 0.317          & 0.504           & 0.319          & 0.508           & 0.323           & 0.512           & 0.324           & 0.513           & -               & -               \\
                                  
                                  & {\model}        & \textbf{0.288} & \textbf{0.459}  & \textbf{0.311} & \textbf{0.492}  & \textbf{0.318} & \textbf{0.504}  & \textbf{0.322} & \textbf{0.509}  & \textbf{0.323}  & 0.510           & \textbf{0.324}  & 0.512           & 0.324           & 0.514           \\ \midrule
\multirow{10}{*}{\textbf{PairRE}} & \textit{Method} & \textit{MRR}   & \textit{Hit@10} & \textit{MRR}   & \textit{Hit@10} & \textit{MRR}   & \textit{Hit@10} & \textit{MRR}   & \textit{Hit@10} & \textit{MRR}    & \textit{Hit@10} & \textit{MRR}    & \textit{Hit@10} & \textit{MRR}    & \textit{Hit@10} \\
                                  & DT              & 0.182          & 0.314           & 0.243          & 0.395           & 0.284          & 0.452           & 0.307          & 0.476           & 0.319           & 0.505           & 0.328           & 0.518           & \textbf{0.332}  & \textbf{0.522}  \\
                                  & Ext             & 0.148          & 0.222           & 0.177          & 0.289           & 0.217          & 0.353           & 0.259          & 0.416           & 0.294           & 0.469           & 0.321           & 0.506           & 0.332           & 0.522           \\
                                  & Ext-L           & 0.150          & 0.249           & 0.196          & 0.294           & 0.219          & 0.333           & 0.271          & 0.436           & 0.309           & 0.489           & 0.326           & 0.513           & 0.332           & 0.522           \\
                                  & Ext-V           & 0.176          & 0.277           & 0.192          & 0.303           & 0.229          & 0.374           & 0.279          & 0.450           & 0.311           & 0.490           & 0.329           & 0.513           & 0.332           & 0.522           \\
                                  & BKD             & 0.198          & 0.332           & 0.251          & 0.407           & 0.288          & 0.453           & 0.311          & 0.487           & 0.321           & 0.508           & 0.330           & 0.521           & -               & -               \\
                                  & TA              & 0.208          & 0.346           & 0.263          & 0.430           & 0.292          & 0.455           & 0.314          & 0.493           & 0.323           & 0.509           & 0.332           & 0.521           & -               & -               \\
                                  & DualDE          & 0.207          & 0.342           & 0.261          & 0.427           & 0.293          & 0.456           & \textbf{0.316} & 0.495           & \textbf{0.326}  & \textbf{0.512}  & \textbf{0.334}  & \textbf{0.524}  & -               & -               \\
                                  & IterDE          & 0.205          & 0.340           & 0.264          & 0.431           & 0.293          & 0.462           & 0.314          & 0.494           & 0.324           & 0.508           & 0.332           & 0.522           & -               & -               \\
                                  & {\model}        & \textbf{0.239} & \textbf{0.384}  & \textbf{0.274} & \textbf{0.437}  & \textbf{0.303} & \textbf{0.466}  & 0.314          & \textbf{0.495}  & 0.324           & 0.510           & 0.329           & 0.521           & 0.330           & 0.520          \\
                                  \bottomrule
\end{tabular}
}
\end{center}
\caption{MRR and Hit@10 of some representative dimensions on FB15K237.}
\label{table_fb_mrr_h1}
\end{table*}

\setlength\tabcolsep{4pt}
\renewcommand\arraystretch{.9}
\begin{table*}[htp]
\begin{center}
    \resizebox{1.0\textwidth}{!}{
\begin{tabular}{cl|cc|cc|cc|cc|cc|cc|cc}
\toprule
\multicolumn{2}{l}{}                                & \multicolumn{2}{c}{\textbf{10d}} & \multicolumn{2}{c}{\textbf{20d}} & \multicolumn{2}{c}{\textbf{40d}} & \multicolumn{2}{c}{\textbf{80d}} & \multicolumn{2}{c}{\textbf{160d}} & \multicolumn{2}{c}{\textbf{320d}} & \multicolumn{2}{c}{\textbf{640d}} \\
\midrule
\multirow{10}{*}{\textbf{TransE}} & \textit{Method} & \textit{Hit@3}  & \textit{Hit@1} & \textit{Hit@3}  & \textit{Hit@1} & \textit{Hit@3}  & \textit{Hit@1} & \textit{Hit@3}  & \textit{Hit@1} & \textit{Hit@3}  & \textit{Hit@1}  & \textit{Hit@3}  & \textit{Hit@1}  & \textit{Hit@3}  & \textit{Hit@1}  \\
                                  & DT              & 0.169           & 0.102          & 0.301           & 0.190          & 0.327           & 0.212          & 0.340           & 0.218          & 0.348           & 0.222           & 0.353           & 0.224           & \textbf{0.358}  & \textbf{0.228}  \\
                                  & Ext             & 0.123           & 0.065          & 0.211           & 0.122          & 0.264           & 0.156          & 0.296           & 0.180          & 0.320           & 0.197           & 0.331           & 0.208           & 0.358           & 0.228           \\
                                  & Ext-L           & 0.118           & 0.065          & 0.192           & 0.115          & 0.256           & 0.157          & 0.292           & 0.180          & 0.316           & 0.198           & 0.333           & 0.210           & 0.358           & 0.228           \\
                                  & Ext-V           & 0.150           & 0.081          & 0.222           & 0.126          & 0.265           & 0.156          & 0.301           & 0.185          & 0.325           & 0.205           & 0.341           & 0.217           & 0.358           & 0.228           \\
                                  & BKD             & 0.178           & 0.106          & 0.308           & 0.198          & 0.336           & 0.208          & 0.349           & 0.222          & 0.349           & 0.223           & 0.354           & 0.226           & -               & -               \\
                                  & TA              & 0.188           & 0.112          & 0.307           & 0.200          & 0.336           & 0.212          & 0.348           & 0.220          & 0.353           & 0.225           & 0.355           & 0.223           & -               & -               \\
                                  & DualDE          & 0.193           & 0.115          & 0.307           & \textbf{0.201} & 0.337           & 0.216          & \textbf{0.351}  & 0.223          & \textbf{0.354}  & 0.226           & 0.356           & 0.227           & -               & -               \\
                                  & IterDE          & 0.187           & 0.112          & 0.299           & 0.185          & 0.333           & 0.214          & 0.351           & 0.222          & 0.353           & 0.223           & 0.354           & 0.224           & -               & -               \\
                                  & {\model}        & \textbf{0.215}  & \textbf{0.122} & \textbf{0.321}  & 0.199          & \textbf{0.338}  & \textbf{0.218} & 0.347           & \textbf{0.223} & 0.351           & \textbf{0.226}  & \textbf{0.356}  & \textbf{0.227}  & 0.358           & 0.227           \\
                                  \midrule
\multirow{10}{*}{\textbf{SimplE}} & \textit{Method} & \textit{Hit@3}  & \textit{Hit@1} & \textit{Hit@3}  & \textit{Hit@1} & \textit{Hit@3}  & \textit{Hit@1} & \textit{Hit@3}  & \textit{Hit@1} & \textit{Hit@3}  & \textit{Hit@1}  & \textit{Hit@3}  & \textit{Hit@1}  & \textit{Hit@3}  & \textit{Hit@1}  \\
                                  & DT              & 0.103           & 0.055          & 0.193           & 0.105          & 0.256           & 0.161          & 0.297           & 0.191          & 0.314           & 0.197           & 0.323           & 0.208           & \textbf{0.324}  & \textbf{0.211}  \\
                                  & Ext             & 0.039           & 0.019          & 0.071           & 0.047          & 0.091           & 0.057          & 0.171           & 0.109          & 0.251           & 0.159           & 0.294           & 0.187           & 0.324           & 0.211           \\
                                  & Ext-L           & 0.043           & 0.035          & 0.048           & 0.037          & 0.111           & 0.040          & 0.131           & 0.093          & 0.216           & 0.134           & 0.281           & 0.177           & 0.324           & 0.211           \\
                                  & Ext-V           & 0.047           & 0.036          & 0.074           & 0.043          & 0.097           & 0.077          & 0.145           & 0.109          & 0.248           & 0.156           & 0.289           & 0.189           & 0.324           & 0.211           \\
                                  & BKD             & 0.123           & 0.064          & 0.201           & 0.115          & 0.261           & 0.164          & 0.299           & 0.191          & 0.308           & 0.202           & 0.318           & \textbf{0.213}  & -               & -               \\
                                  & TA              & 0.133           & 0.073          & 0.210           & 0.123          & 0.276           & 0.175          & 0.302           & 0.195          & 0.318           & 0.203           & 0.323           & 0.211           & -               & -               \\
                                  & DualDE          & 0.130           & 0.071          & 0.224           & 0.115          & 0.279           & 0.175          & 0.305           & 0.196          & 0.324           & \textbf{0.208}  & \textbf{0.326}  & 0.211           & -               & -               \\
                                  & IterDE          & 0.132           & 0.069          & 0.217           & 0.118          & 0.276           & 0.174          & 0.303           & 0.192          & \textbf{0.326}  & 0.204           & 0.324           & 0.212           & -               & -               \\
                                  & {\model}        & \textbf{0.164}  & \textbf{0.073} & \textbf{0.254}  & \textbf{0.158} & \textbf{0.288}  & \textbf{0.177} & \textbf{0.305}  & \textbf{0.196} & 0.319           & 0.205           & 0.318           & 0.209           & 0.322           & 0.209           \\
                                  \midrule
\multirow{10}{*}{\textbf{RotatE}} & \textit{Method} & \textit{Hit@3}  & \textit{Hit@1} & \textit{Hit@3}  & \textit{Hit@1} & \textit{Hit@3}  & \textit{Hit@1} & \textit{Hit@3}  & \textit{Hit@1} & \textit{Hit@3}  & \textit{Hit@1}  & \textit{Hit@3}  & \textit{Hit@1}  & \textit{Hit@3}  & \textit{Hit@1}  \\
                                  & DT              & 0.284           & 0.168          & 0.330           & 0.207          & 0.346           & 0.223          & 0.352           & 0.224          & 0.353           & 0.229           & 0.357           & 0.230           & \textbf{0.363}  & \textbf{0.234}  \\
                                  & Ext             & 0.152           & 0.080          & 0.225           & 0.129          & 0.278           & 0.170          & 0.304           & 0.190          & 0.322           & 0.203           & 0.335           & 0.217           & 0.363           & 0.234           \\
                                  & Ext-L           & 0.147           & 0.078          & 0.209           & 0.121          & 0.247           & 0.146          & 0.275           & 0.166          & 0.312           & 0.193           & 0.333           & 0.209           & 0.363           & 0.234           \\
                                  & Ext-V           & 0.174           & 0.097          & 0.218           & 0.126          & 0.264           & 0.159          & 0.293           & 0.182          & 0.319           & 0.201           & 0.336           & 0.213           & 0.363           & 0.234           \\
                                  & BKD             & 0.306           & 0.193          & 0.338           & 0.214          & 0.352           & 0.224          & 0.354           & 0.230          & 0.356           & 0.230           & 0.358           & 0.231           & -               & -               \\
                                  & TA              & 0.308           & 0.196          & 0.339           & 0.216          & 0.353           & 0.225          & 0.358           & 0.229          & 0.359           & 0.229           & 0.358           & 0.231           & -               & -               \\
                                  & DualDE          & 0.311           & 0.197          & 0.341           & 0.216          & 0.353           & \textbf{0.227} & \textbf{0.360}  & 0.230          & 0.361           & 0.232           & 0.361           & 0.233           & -               & -               \\
                                  & IterDE          & 0.307           & 0.195          & 0.342           & 0.215          & 0.355           & 0.225          & 0.359           & \textbf{0.232} & \textbf{0.363}  & 0.233           & 0.362           & \textbf{0.234}  & -               & -               \\
                                  & {\model}        & \textbf{0.324}  & \textbf{0.201} & \textbf{0.344}  & \textbf{0.216} & \textbf{0.355}  & 0.225          & 0.357           & 0.231          & 0.358           & \textbf{0.233}  & \textbf{0.362}  & 0.233           & 0.362           & 0.232           \\
                                  \midrule
\multirow{10}{*}{\textbf{PairRE}} & \textit{Method} & \textit{Hit@3}  & \textit{Hit@1} & \textit{Hit@3}  & \textit{Hit@1} & \textit{Hit@3}  & \textit{Hit@1} & \textit{Hit@3}  & \textit{Hit@1} & \textit{Hit@3}  & \textit{Hit@1}  & \textit{Hit@3}  & \textit{Hit@1}  & \textit{Hit@3}  & \textit{Hit@1}  \\
                                  & DT              & 0.198           & 0.116          & 0.262           & 0.162          & 0.312           & 0.202          & 0.337           & 0.222          & 0.352           & 0.227           & 0.364           & 0.235           & \textbf{0.368}  & \textbf{0.237}  \\
                                  & Ext             & 0.158           & 0.107          & 0.187           & 0.118          & 0.236           & 0.149          & 0.283           & 0.182          & 0.325           & 0.207           & 0.354           & 0.230           & 0.368           & 0.237           \\
                                  & Ext-L           & 0.159           & 0.099          & 0.196           & 0.134          & 0.238           & 0.159          & 0.298           & 0.188          & 0.342           & 0.219           & 0.359           & 0.233           & 0.368           & 0.237           \\
                                  & Ext-V           & 0.181           & 0.116          & 0.192           & 0.125          & 0.250           & 0.154          & 0.307           & 0.193          & 0.343           & 0.221           & 0.362           & 0.237           & 0.368           & 0.237           \\
                                  & BKD             & 0.215           & 0.132          & 0.265           & 0.168          & 0.314           & 0.203          & 0.343           & 0.224          & 0.355           & 0.233           & 0.366           & 0.236           & -               & -               \\
                                  & TA              & 0.226           & 0.139          & 0.291           & 0.182          & 0.316           & 0.210          & 0.347           & 0.224          & 0.358           & 0.232           & 0.368           & 0.235           & -               & -               \\
                                  & DualDE          & 0.224           & 0.139          & 0.286           & 0.179          & 0.318           & 0.212          & 0.351           & 0.226          & \textbf{0.359}  & \textbf{0.234}  & \textbf{0.371}  & \textbf{0.238}  & -               & -               \\
                                  & IterDE          & 0.225           & 0.135          & 0.293           & 0.185          & 0.324           & 0.212          & \textbf{0.352}  & 0.224          & 0.357           & 0.234           & 0.369           & 0.236           & -               & -               \\
                                  & {\model}        & \textbf{0.253}  & \textbf{0.172} & \textbf{0.299}  & \textbf{0.189} & \textbf{0.327}  & \textbf{0.213} & 0.346           & \textbf{0.224} & 0.357           & 0.232           & 0.366           & 0.236           & 0.368           & 0.235          \\
                                  \bottomrule
\end{tabular}
}
\end{center}
\caption{Hit@3 and Hit@1 of some representative dimensions on FB15K237.}
\label{table_fb_h10_h3}
\end{table*}

\section{Ablation Study}
\label{sec:exper_ablation}
We conduct ablation studies to evaluate the effect of three modules in {\model}: the mutual learning mechanism (MLM), the evolutionary improvement mechanism (EIM), and the dynamic loss weight (DLW). Table~\ref{table_ablation} shows the MRR and Hit@$k$ ($k=1,3,10$) of {\model} removing these modules respectively on WN18RR and TransE.

\begin{table*}[htbp]\setlength\tabcolsep{4pt}

\begin{center}
    \resizebox{0.96\textwidth}{!}{
\begin{tabular}{l|cccc|cccc|cccc|cccc}
\toprule
\multirow{2}{*}{dim} & \multicolumn{4}{c|}{{\model}}                                       & \multicolumn{4}{c|}{{\model} w/o MLM}                                   & \multicolumn{4}{c|}{{\model} w/o EIM}                                   & \multicolumn{4}{c}{{\model} w/o DLW}                                   \\
                     & \textit{MRR}   & \textit{Hit@10} & \textit{Hit@3} & \textit{Hit@1} & \textit{MRR}   & \textit{Hit@10} & \textit{Hit@3} & \textit{Hit@1} & \textit{MRR}   & \textit{Hit@10} & \textit{Hit@3} & \textit{Hit@1} & \textit{MRR}   & \textit{Hit@10} & \textit{Hit@3} & \textit{Hit@1} \\
\midrule
10                   & .170          & \textbf{.388}           & \textbf{.269} & .036          & .149          & .335           & .234          & .032          & .169          & .388  & .267          & \textbf{.037} & \textbf{.171} & .387           &.268 & .035          \\
20                   & \textbf{.219} & \textbf{.491}  & \textbf{.369} & .042          & .197          & .437           & .323          & .032          & .217          & .488           & .366          & \textbf{.044} & .218          & .487           & .367          & .039          \\
40                   & \textbf{.232} & \textbf{.518}  & .399          & \textbf{.048} & .224          & .496           & .379          & .029          & .232          & .517           & \textbf{.403} & .042          & .232          & .517           & .402          & .037          \\
80                   & .232          & .523           & .404          & \textbf{.042} & .228          & .521           & .399          & .033          & \textbf{.235} & \textbf{.529}  & .408          & .037          & .234          & .523           & \textbf{.410} & .041          \\
160                  & \textbf{.236} & \textbf{.529}  & \textbf{.407} & \textbf{.037} & .234          & .525           & .406          & .034          & .234          & .527           & .405          & .032          & .235          & .527           & .405          & .032          \\
320                  & \textbf{.237} & \textbf{.536}  & \textbf{.410} & .033          & .236          & .532           & .409          & \textbf{.035} & .233          & .530           & .398          & .031          & .234          & .533           & .405          & .029          \\
640                  & .237          & \textbf{.537}  & \textbf{.412} & .031          & \textbf{.238} & .535           & .412          & \textbf{.042} & .232          & .528           & .402          & .029          & .233          & .530           & .396          & .025  \\
\bottomrule
\end{tabular}
}
\end{center}
\caption{Ablation study on WN18RR with TransE.}\label{table_ablation}
\end{table*}

\subsection{Mutual Learning Mechanism (MLM)}
We remove the mutual learning mechanism from {\model} and keep the other parts unchanged, where \eqref{equ:loss} is rewritten as
\begin{equation}
    L = \sum_{i=1}^n \exp \left(\frac{w_3 \cdot d_i}{d_{n}}\right) \cdot L_{EI}^i. 
    \label{equ:loss_re_MLM}
\end{equation}

From the result of ``{\model} w/o MLM'' in Table \ref{table_ablation}, we find that after removing the mutual learning mechanism, the performance of low-dimensional sub-models deteriorates seriously since the low-dimensional sub-models can not learn from the high-dimensional sub-models. For example, the MRR of the 10-dimensional sub-model decreased by $12.4\%$, and the MRR of the 20-dimensional sub-model decreased by $10\%$. While the performance degradation of the high-dimensional sub-model is not particularly obvious, and the MRR of the highest-dimensional sub-model ($dim=640$) is not worse than that of {\model}, which is because to a certain degree, removing the mutual learning mechanism also avoids the negative influence to high-dimensional sub-models from low-dimensional sub-models. On the whole, this mechanism greatly improves the performance of low-dimensional sub-models.

\subsection{Evolutionary Improvement Mechanism (EIM)}
In this part, we replace evolutionary improvement loss $L_{EI}^i$ in \eqref{equ:loss} with the regular KGE loss $L_{KGE}^i$:
\begin{equation}
\begin{aligned}
    L_{KGE}^i = \sum_{(h,r,t)\in \mathcal{T}\cup \mathcal{T}^-}  y \log \sigma(s_{(h, r, t)}^i) \\ + (1-y)\log (1-\sigma(s_{(h, r, t)}^i)).
\end{aligned}
    \label{equ:loss_re_EIM}
\end{equation}

From the result of ``{\model} w/o EIM'' in Table \ref{table_ablation}, we find that removing the evolutionary improvement mechanism mainly degrades the performance of high-dimensional sub-models. 
While due to the existence of the mutual learning mechanism, the low-dimensional sub-model can still learn from the high-dimensional sub-model, so as to ensure the certain performance of the low-dimensional sub-model. In addition, we also find that as the dimension increases to a certain extent, the performance of the sub-model does not improve, and even begins to decline. We guess that this is because the mutual learning mechanism makes every pair of neighbor sub-models learn from each other, resulting in some low-quality or wrong knowledge gradually transferring from the low-dimensional sub-models to the high-dimensional sub-models, and when the evolutionary improvement mechanism is removed, the high-dimensional sub-models can no longer correct the wrong information from the low-dimensional sub-models. The higher the dimension of the sub-model, the more the accumulated error, so the performance of the high-dimensional sub-models is seriously damaged.
On the whole, this mechanism mainly helps to improve the effect of high-dimensional sub-models.

\subsection{Dynamic Loss Weight (DLW)}
To study the effect of the dynamic loss weight, we fix the ratio of all mutual learning losses to all evolutionary improvement losses as $1:1$, and \eqref{equ:loss} is rewritten as
\begin{equation}
    L = \sum_{i=2}^n L_{ML}^{i-1,i} + \sum_{i=1}^n L_{EI}^i.
    \label{equ:loss_re_DLW}
\end{equation}

According to the result of ``{\model} w/o DLW'' in Table \ref{table_ablation}, the overall results of ``{\model} w/o DLW'' are in the middle of the results of ``{\model} w/o MLM'' and ``{\model} w/o EIM'': the performance of the low-dimensional sub-model is better than that of ``{\model} w/o MLM'', and the performance of the high-dimensional sub-model is better than that of ``{\model} w/o EIM''. On the whole, its results are more similar to ``{\model} w/o EIM'', that is, the performance of the low-dimensional sub-model does not change much, while the performance of the high-dimensional sub-model decreases more significantly. We believe that for the high-dimensional sub-model, the proportion of mutual learning loss is still too large, which makes it more negatively affected by the low-dimensional sub-model. This result indicates that the dynamic loss weight plays a role in adaptively balancing multiple losses and contributes to improving overall performance.

\section{Details of applying the trained KGE by {\model} to real applications}
\label{sec:detail_application}
The SKG is used in many tasks related to users, and injecting user embeddings trained over SKG into downstream task models is a common and practical way.

User labeling is one of the common user management tasks that e-commerce platforms run on backend servers. We model user labeling as a multiclass classification task for user embeddings with a 2-layer MLP:
\begin{equation}
\label{equ_usercls}
    \mathcal{L}=-\frac{1}{|\mathcal{U}|}\sum_{i=1}^{|\mathcal{U}|}\sum_{j=1}^{|\mathcal{CLS}|} y_{ij} \log(\mathrm{MLP}(u_i)),
\end{equation}
where $u_{i}$ is the $i$-th user's embedding, the label $y_{ij} = 1$ if user $u_{i}$ belongs to class $cls_j$, otherwise $y_{ij} = 0$. 

The product recommendation task is to properly recommend items to users that users will interact with a high probability and it often runs on terminal devices. Following PKGM~\cite{DBLP:conf/icde/ZhangWYWZC21}, which recommends items to users using the neural collaborative filtering (NCF)~\cite{DBLP:conf/www/HeLZNHC17} framework with the help of pre-trained user embeddings as service vectors, we add trained user embeddings over SKG as service vectors to NCF. In NCF, the MLP layer is used to learn item-user interactions based on the latent feature of the user and item, that is, for a given user-item pair $user_i-item_j$, the interaction function is 
\begin{equation}
    \phi_{1}^{M L P}\left(p_{i}, q_{j}\right)=\mathrm{MLP}([p_{i}; q_{j}]),
\label{equ_item_recom_ori}
\end{equation}where $p_i$ and $q_j$ are latent feature vectors of user and item learned in NCF. We add the trained user embedding $u_i$ to NCF's MLP layer and rewrite Equation~\eqref{equ_item_recom_ori} as
\begin{equation}
    \phi_{1}^{M L P}\left(p_{i}, q_{j}, u_i\right)=\mathrm{MLP}([p_{i}; q_{j}; u_i]),
\label{equ_item_recom}
\end{equation}
 and the other parts of NCF stay the same as in PKGM~\cite{DBLP:conf/icde/ZhangWYWZC21}.

We train entity and relation embeddings for SKG based on TransE~\cite{DBLP:conf/nips/BordesUGWY13} and input the trained entity (user) embedding into Equation~\eqref{equ_usercls} and Equation~\eqref{equ_item_recom}. 

\begin{figure*}[htbp]
\includegraphics[width=.9\linewidth]{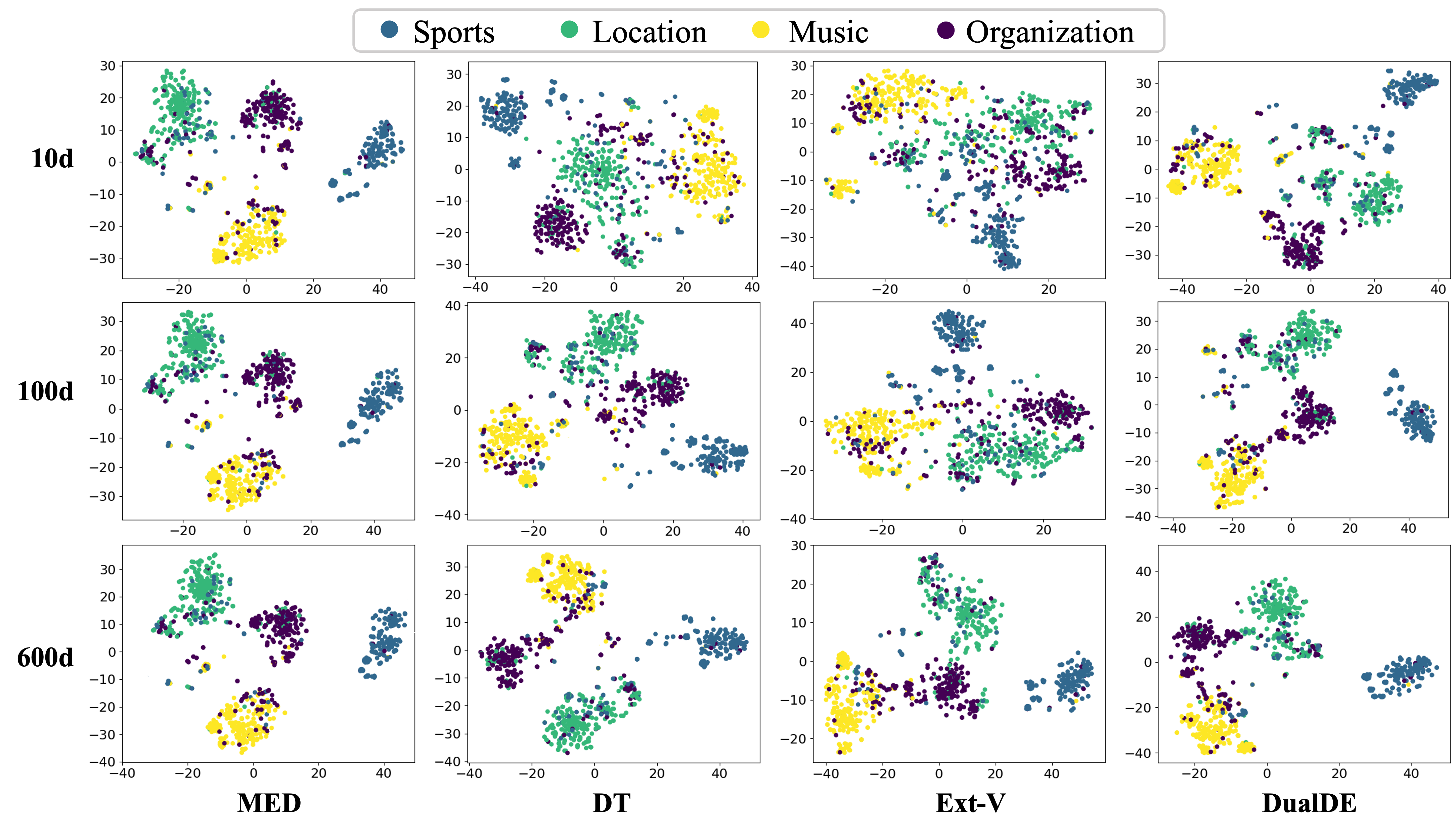}
    \caption{Clustering on FB15K237 with RotatE.}
    \label{fig:visual}
\end{figure*}
\section{Details of extending {\model} to language model BERT-base}
\label{sec:detail_bert}
\subsection{Dataset and Evaluation Metric}
For the experiments extending {\model} to BERT, we adopt the common GLUE \cite{DBLP:conf/iclr/WangSMHLB19} benchmark for evaluation. To be specific, we use the development set of the GLUE benchmark which includes four tasks: Paraphrase Similarity Matching, Sentiment Classification, Natural Language Inference, and Linguistic Acceptability. For Paraphrase Similarity Matching, we use MRPC \cite{DBLP:conf/acl-iwp/DolanB05}, QQP and STS-B \cite{DBLP:conf/lrec/ConneauK18} for evaluation. For Sentiment Classification, we use SST-2 \cite{DBLP:conf/emnlp/SocherPWCMNP13}. For Natural Language Inference, we use MNLI \cite{DBLP:conf/naacl/WilliamsNB18}, QNLI \cite{DBLP:conf/emnlp/RajpurkarZLL16}, and RTE for evaluation. In terms of evaluation metrics, we follow previous work \cite{DBLP:conf/naacl/DevlinCLT19,DBLP:conf/emnlp/SunCGL19}. For MRPC and QQP, we report F1 and accuracy. For STS-B, we consider Pearson and Spearman correlation as our metrics. The other tasks use accuracy as the metric. For MNLI, the results of MNLI-m and MNLI-mm are both reported separately.
\subsection{Baselines}
For comparison, we choose Knowledge Distillation (KD) models and Hardware-Aware Transformers \cite{DBLP:conf/acl/WangWLCZGH20} (HAT) customized for transformers as baselines. For the KD models, we compare {\model} with Basic KD (BKD) \cite{DBLP:journals/corr/HintonVD15}, Patient KD (PKD) \cite{DBLP:conf/emnlp/SunCGL19},  Relational Knowledge Distillation (RKD) \cite{DBLP:conf/cvpr/ParkKLC19}, Deep Self-attention Distillation (MiniLM) \cite{DBLP:conf/nips/WangW0B0020}, Meta Learning-based KD (MetaDistill) \cite{DBLP:conf/acl/ZhouXM22} and Feature Structure Distillation (FSD) \cite{DBLP:journals/eswa/JungKNK23}. For the comparability of the results, we choose 4-layer BERT (BERT$_4$) or 6-layer BERT (BERT$_6$) as the student model architectures, which guarantees that the number of model parameters ($\textit{\#P(M)}$) or $\textit{speedup}$ is comparable. For HAT, we use the same model architecture as our {\model} for training and show the results of sub-models with three parameter scales.
\subsection{Implementation}
To implement {\model} on BERT, for the word embedding layer, all sub-models share the front portion of embedding parameters in the same way as in KGE, and for the transformer layer, all sub-models share the front portion of weight parameters as in HAT~\cite{DBLP:conf/acl/WangWLCZGH20}. Specifically, assuming that the embedding dimension of the largest BERT model $B_{n}$ is $d_{n}$, and the embedding dimension of the sub-model $B_i$ is $d_i$, for any parameter matrix with the shape $x
\times y$ in $B_n$, the front portion sub-matrix of it with the shape $\frac{d_i}{d_n}x \times \frac{d_i}{d_n}y$ is the parameter matrix of the corresponding position in $B_i$. Finally, it just need to replace the triple score $s_{(h,r,t)}$ in
Equation~\eqref{equ:Loss_ML}, Equation~\eqref{equ:pos_w}, Equation~\eqref{equ:neg_w}, and Equation~\eqref{equ:loss_EI} with the logits output for the corresponding category of the classifier in the classification task. 

We set $n=4$ for BERT applying {\model}, and 4 sub-models have the following settings: [768, 512, 256, 128] for embedding dim and [768, 512, 256, 128]  for hidden dim, [12, 12, 6, 6] for the head number in attention modules, 12 for encoder layer number. 

\section{Visual analysis of embedding}
\label{sec:exper_visual}
We select four primary entity categories (`organization', `sports', `location', and `music') that contain more than 300 entities in FB15K237, and randomly select 250 entities for each. We cluster these entities' embeddings of 3 different dimensions ($d$=10, 100, 600) by the t-SNE algorithm, and the clustering results are visualized in Fig.~\ref{fig:visual}.
Under the same dimension, the clustering result of {\model} is always the best, followed by DualDE, while the result of Ext-V is generally poor, which is consistent with the conclusion in Section~\ref{sec:exper_compare}. 
We also find some special phenomenons for {\model} when dimension increases: 1) the nodes of the `sports' gradually become two clusters meaning {\model} learns more fine-grained category information as dimension increases, and 2) the relative distribution among different categories hardly changes and shows a trend of ``inheritance''  and ``improvement''. This further proves {\model} achieves our expectation that high-dimensional sub-models retain the ability of low-dimensional sub-models, and can learn more knowledge than low-dimensional sub-models.

\end{document}